%% file: main.tex
\documentclass[10pt,letterpaper,twocolumn]{article}
\usepackage{arxiv}
\usepackage[T1]{fontenc}
\usepackage[utf8]{inputenc}
\usepackage[hyphens]{url}
\usepackage{graphicx}
\usepackage[authoryear,round]{natbib}
\usepackage{bibunits}
\usepackage{caption}
\usepackage{cuted}
\usepackage{booktabs,longtable,tabularx,diagbox}
\usepackage{amsmath,amssymb}
\usepackage[table]{xcolor}
\usepackage{multirow,array,microtype,placeins}
\usepackage[hidelinks]{hyperref}
\renewcommand{\dbltopfraction}{0.9}
\renewcommand{\headeright}{}
\renewcommand{\undertitle}{A Preprint}
\renewcommand{\shorttitle}{A Scaling Study for fMRI Foundation Models}
\renewenvironment{abstract}{%
  \par\small
  \begin{center}\large\bfseries\scshape Abstract\end{center}%
  \noindent\ignorespaces
}{\par\medskip}
\hypersetup{
  pdftitle={A Scaling Study for fMRI Foundation Models},
  pdfauthor={Wenhao Ye, Xuanye Pan, Junfeng Xia, Junxiang Zhang, Mo Wang, Quanying Liu}
}
\title{A Scaling Study for fMRI Foundation Models}
\author{%
  Wenhao Ye\textsuperscript{1,2}\quad
  Xuanye Pan\textsuperscript{1}\quad
  Junfeng Xia\textsuperscript{1}\quad
  Junxiang Zhang\textsuperscript{1,4}\\[3pt]
  \bfseries Mo Wang\textsuperscript{1,3,\textdagger}\quad
  Quanying Liu\textsuperscript{1,\textdagger}\\[6pt]
  {\normalfont\small\textsuperscript{1}Southern University of Science and Technology\quad
  \textsuperscript{2}Shenzhen University}\\
  {\normalfont\small\textsuperscript{3}University of Warwick\quad
  \textsuperscript{4}SLAI}\\[2pt]
  {\normalfont\small\textsuperscript{\textdagger}Equal author}
}
\date{}

\definecolor{panelgray}{HTML}{F0F2F5}
\definecolor{panelline}{HTML}{9AA3AF}

\newcommand{\best}[1]{\textbf{#1}}
\newcommand{\neurojepa}{NeuroJEPA}

\begin{document}
\twocolumn[\maketitle]
\thispagestyle{empty}
\input{sections/00_abstract}
\input{sections/00b_teaser}
\input{sections/01_introduction}
\input{sections/02_related_work}
\input{sections/03_objective}
\input{sections/04_protocol}
\input{sections/05_scaling}
\input{sections/06_final_evaluation}
\input{sections/07_discussion}
\FloatBarrier
\clearpage
\bibliographystyle{plainnat}
\bibliography{references}
\clearpage
\appendix
\input{sections/08_appendix}
\begin{bibunit}[plainnat]
  \input{sections/09_baselines}
  \onecolumn
  \input{sections/09_datasets}
  \clearpage
  \twocolumn
  \begingroup
    \renewcommand{\refname}{Appendix References}
    \putbib[appendix_references]
  \endgroup
\end{bibunit}
\end{document}

%% file: sections/00_abstract.tex
\begin{abstract}
Scaling laws have guided large-model development in computer vision and natural language processing, but the relationships among data, model size, and compute remain unclear for functional magnetic resonance imaging (fMRI) foundation models. Here, we conduct a controlled empirical study using pretraining data from more than 200 source datasets and over 10,000 GPU-hours of experiments. Holding the pretraining framework and downstream protocol fixed, we vary pretraining data size, model size, and training duration.
Downstream performance generally improves with compute, yet models using similar compute can perform substantially differently. Additional pretraining data bring larger gains at larger model sizes, suggesting that data and model size should be scaled together. At matched compute, increasing pretraining data benefits more tasks than increasing model size, although the pattern varies across tasks.
We then use in-distribution (ID) downstream performance to select the combination of pretraining data size, model size, and training duration at two fixed compute budgets. The resulting models are locked before out-of-distribution (OOD) evaluation. They achieve the highest average performance across the evaluated OOD tasks among the compared fMRI foundation models while using less pretraining compute. Overall, our results show that compute alone does not characterize fMRI scaling: performance depends on how pretraining data, model size, and training duration are combined.
Code is available at \href{https://github.com/derrz2/neurojepa}{\textcolor[HTML]{1F4E79}{link}}.
\end{abstract}

%% file: sections/00b_teaser.tex
\begin{figure}[!t]
    \centering
    \includegraphics[width=\linewidth]{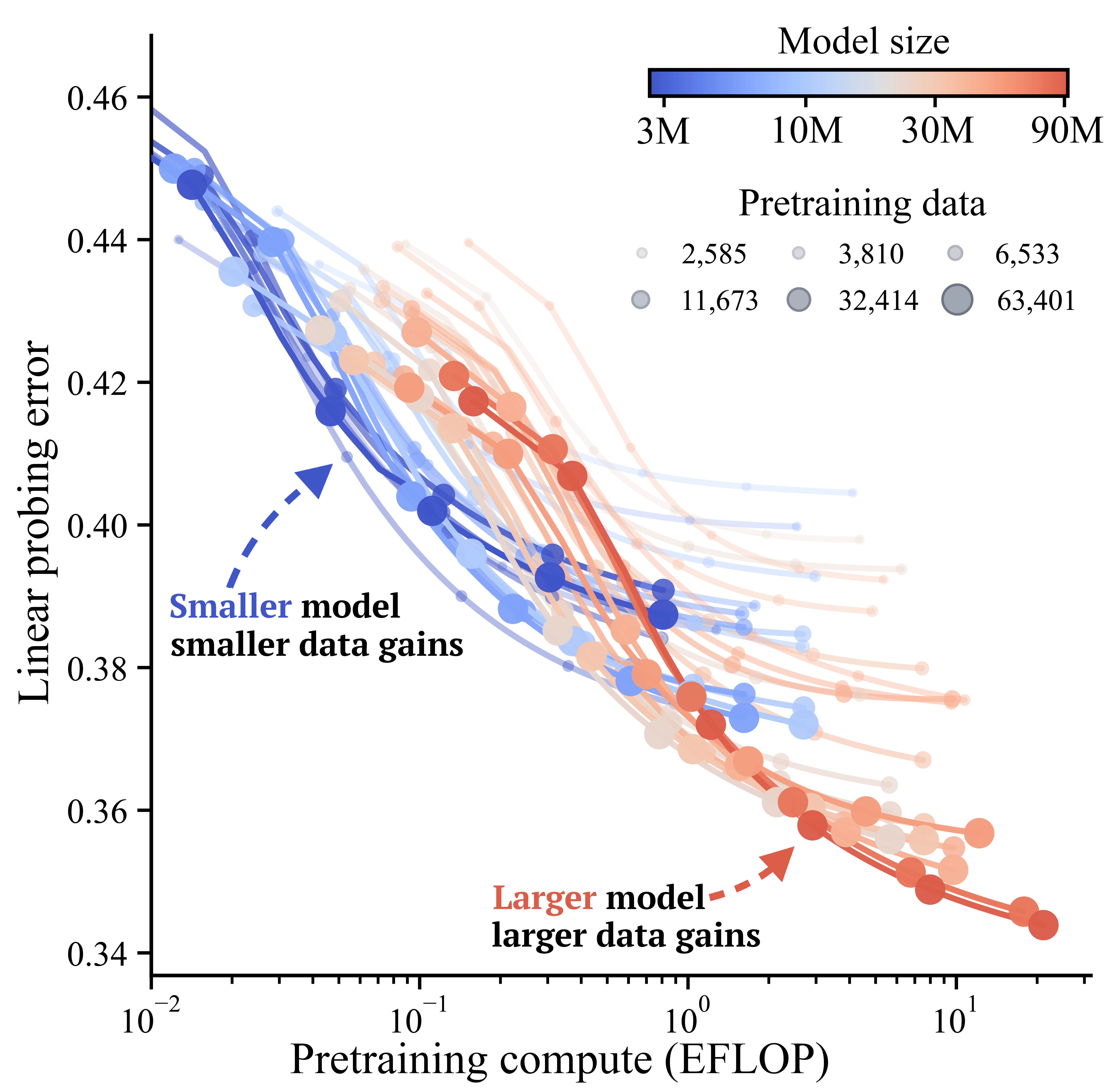}
    \caption{\textbf{Compute allocation shapes downstream performance.}
    Each trajectory follows one data-model configuration; 
    marker area denotes pretraining data size and color denotes model size. 
    At comparable compute, configurations reach different mean linear-probe errors across 12 tasks; 
    within the evaluated range, data gains are larger for larger models.}
    \label{fig:error_landscape}
\end{figure}

%% file: sections/01_introduction.tex
\section{Introduction}

Scaling studies in language and vision have provided practical guidance for
relating data, model size, and compute to performance
\citep{kaplan_scaling_laws,chinchilla,scaling_vit,openclip_scaling}. Functional
magnetic resonance imaging (fMRI) operates under a different constraint:
pretraining relies on a finite collection of human recordings, while labeled
downstream datasets are often small
\citep{reproducibility_neuroimaging,brainlm,brain_semantoks,slim_brain,brainworld}.
More pretraining data, a larger model, and continued training can consume
similar compute without necessarily producing the same downstream benefit.
The central question is therefore how pretraining data size, model size, and training progress jointly shape downstream performance.

Existing fMRI foundation-model studies provide encouraging evidence that
larger pretraining datasets or larger models can improve downstream
performance
\citep{brainlm,brain_semantoks,braingfm,brainmass,brainharmonix,omnifmri}.
However,
these factors have largely been examined separately or only at the end of
training. Consequently, it remains unclear whether additional data become
more useful as models grow, whether this relationship changes during training,
and whether different data--model configurations remain comparable when they
use similar compute. We address these questions through a controlled study
centered on downstream performance.

To isolate scaling behavior, we hold the pretraining objective, architecture family, and downstream protocol fixed.
All models use \neurojepa{}, a simple ROI-time implementation following
LeJEPA~\citep{lejepa}, as a common pretraining method rather than as a scaling
variable. Our experiments use data assembled from more than 200 source
datasets. 
We follow each supported data--model configuration through multiple stages of training and
evaluate frozen representations on 12 downstream tasks. This design lets us
compare continued training within one configuration with differences across
configurations using similar estimated compute. 

Three patterns emerge. First, downstream performance generally improves with
additional training, but models using similar compute can still perform
differently. Second, additional pretraining data are associated with larger
gains for larger models, and this relationship becomes clearer later in
training. Third, at matched compute, increasing pretraining data provides the
more consistent gain across downstream tasks, whereas the benefit of
increasing model size is more task dependent. These results show
that fMRI scaling cannot be characterized by compute alone: pretraining data
size, model size, and training compute need to be considered jointly. 

Finally, we fit in-distribution (ID) downstream performance over the observed combinations of pretraining data size, model size, and training duration. 
At each of two prespecified compute budgets, the fitted relationship selects the combination with the best predicted ID performance. We lock both selections before examining any out-of-distribution (OOD) results and then compare the selected models with released fMRI foundation models.

We make three contributions:
\begin{itemize}
\setlength{\itemsep}{1pt}
\setlength{\parskip}{0pt}
\setlength{\parsep}{0pt}
\item We conduct a controlled, scaling study of
fMRI foundation models. Using data assembled from more than 200 source
datasets and over 10,000 GPU-hours of experiments, we evaluate six nested
pretraining data sizes, model sizes ranging from approximately 2M to 92M
encoder parameters, and multiple compute budgets on 12 frozen linear-probe
tasks under one fixed protocol.
\item We characterize how pretraining data and model size scale together
throughout training. By following each supported data--model configuration
over time, we show that larger models are associated with greater
gains from additional pretraining data.
\item We compare data scaling and model scaling task by task at matched
compute. Most tasks show a clearer benefit from
additional pretraining data, while the benefit of increasing model size is
more task dependent. We then use ID downstream performance to select two
models under fixed compute budgets and evaluate the locked models on OOD
tasks.
\end{itemize}

%% file: sections/02_related_work.tex
\section{Related Work}

\paragraph{Scaling data, model size, and compute.}
Scaling studies in language and vision vary data and model size over several
orders of magnitude, fit predictable loss or performance trends, and use them
to compare compute-efficient training configurations
\citep{kaplan_scaling_laws,chinchilla,scaling_vit,openclip_scaling}. Although
data constraints are now also recognized in language-model training
\citep{muennighoff_data_constrained}, fMRI begins in a more
acquisition-constrained regime. Additional pretraining data require human
recordings, while reproducible brain--phenotype associations can require
thousands of participants even though typical neuroimaging cohorts are much
smaller \citep{reproducibility_neuroimaging}. Expanding an fMRI corpus also
commonly adds participants, sites, scanners, and acquisition protocols, so its
data axis changes both the amount and composition of the available evidence
\citep{flexibrain,brain_dit}.
Moreover, the relevant outcome is downstream performance across label-limited
tasks, not pretraining loss alone \citep{braintaskonomy}. We therefore study how pretraining data,
model size, and estimated compute jointly relate to downstream performance,
and how these relationships change with training progress and across tasks.

\paragraph{Scaling fMRI pretraining.}
BrainLM introduced explicit data- and model-size experiments into fMRI
foundation modeling, reporting masked-signal reconstruction and limited
downstream comparisons across pretraining scales \citep{brainlm}.
Brain-Semantoks studied data scaling with a fixed architecture across frozen
linear probes \citep{brain_semantoks}. These studies
show favorable trends along individual scaling axes, but do not jointly
characterize how pretraining data, model size, and training progress should be
allocated under a common compute budget. CortexMAE provides the closest
systematic study: it performs data- and model-size sweeps on HCP-YA, fits
power-law trends for masked-reconstruction loss, and reports downstream trends
on four selected targets \citep{cortexmae_scaling}. It also examines
performance across training progress, but does not compare alternative data--model
allocations at matched total compute. We instead relate downstream performance
to nested multi-study pretraining data sizes, model size, and training progress.

\paragraph{Self-supervised objectives for fMRI.}
Early fMRI foundation models primarily learned by reconstructing masked BOLD
signals. BrainLM reconstructs ROI-time patches, while CortexMAE applies masked
reconstruction to cortical flat maps \citep{brainlm,cortexmae_scaling}.
BrainMass is a hybrid masked-modeling and latent-alignment method for
functional-connectivity matrices. It constructs two pseudo-functional
connectivity views by dropping BOLD time points, encodes them with online and
EMA-updated target networks, and aligns their normalized latent embeddings
with an online predictor. This alignment is trained jointly with masked-ROI
identification and reconstruction losses \citep{brainmass}. Brain-Semantoks
learns temporally stable representations through self-distillation between
temporal views. It combines an EMA teacher--student pair with a
functional-network tokenizer, masked-token prediction, coding-rate
regularization, and an early-training curriculum \citep{brain_semantoks}. For
controlled scaling, we use a deliberately simpler pretraining method.
Following LeJEPA~\citep{lejepa}, \neurojepa{} aligns global and local views of
the same ROI-time segment with a shared encoder and applies SIGReg to encourage
isotropic latent geometry. Compared with BrainMass and Brain-Semantoks, it
uses no EMA target network; it also requires no masked reconstruction head,
semantic tokenizer, or training curriculum. We do not claim that this
objective is universally optimal; it provides a fixed pretraining method for
studying how pretraining data, model size, and training progress relate to
downstream performance.

%% file: sections/03_objective.tex
\section{Fixed Pretraining Method}
\label{sec:objective}

This section defines the representation learner used in every scaling
experiment. \neurojepa{} applies the LeJEPA objective~\citep{lejepa} to
ROI time series. The purpose of this section is to make the fixed pretraining method clear before the
scaling variables are introduced.

\begin{figure}[t]
    \centering
    \includegraphics[width=\linewidth]{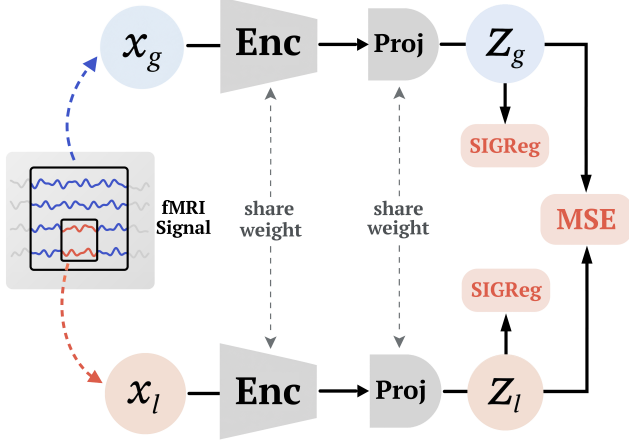}
    \caption{\neurojepa{} pretraining, illustrated with one of two global views
    and one of six local views. All views share the same encoder and projection
    head. The loss aligns every view to the mean global representation and
    applies isotropic geometry regularization to each view batch.}
    \label{fig:neurojepa_pipeline}
\end{figure}

An input segment is a matrix of ROI signals over time. For a minibatch of $B$
segments, let $x_i\in\mathbb{R}^{R\times T}$ denote segment $i$, where $R$
is the number of cortical ROIs ($R=100$ in all experiments) and $T$ is the
number of time points. We construct $V=8$ augmented views of each segment:
two \emph{global views} that retain broad
ROI-time context and six \emph{local views} that retain less context. The
$v$th view of segment $i$ is $x_i^{(v)}$. All views come from the same input
segment; the complete sampling recipe is in
the supplementary material.

The shared encoder $f_\theta$ is a Transformer over small blocks of the
ROI-by-time matrix. It maps each view to one pooled feature vector. A shared
projection head $g_\phi$, a small network applied after the encoder, maps that
feature vector to the
$d_z$-dimensional embedding
$z_i^{(v)}=g_\phi(f_\theta(x_i^{(v)}))\in\mathbb{R}^{d_z}$.
Let $\mathcal{G}$ be the set of global-view indices. The mean global embedding
for segment $i$ is
\[
    c_i=\frac{1}{|\mathcal{G}|}\sum_{v\in\mathcal{G}}z_i^{(v)}.
\]
All global and local embeddings are aligned to this mean:
\[
    \mathcal{L}_{\mathrm{align}}
    =\frac{1}{BV}\sum_{i=1}^{B}\sum_{v=1}^{V}
      \left\|z_i^{(v)}-c_i\right\|_2^2 .
\]
Gradients pass through the mean and every view embedding; no separate target
encoder is used.

The second term is Sketched Isotropic Gaussian Regularization (SIGReg) from
LeJEPA. For one view index $v$, SIGReg projects the batch
$\{z_i^{(v)}\}_{i=1}^{B}$ onto random one-dimensional directions and compares
each projected distribution with a standard Gaussian using the Epps--Pulley
statistic. This discourages collapsed or strongly concentrated embeddings.
Let $\lambda\in(0,1)$ be its loss weight. The complete objective is
\[
    \mathcal{L}_{\mathrm{NeuroJEPA}}
    =
    (1-\lambda)\mathcal{L}_{\mathrm{align}}
    +
    \frac{\lambda}{V}\sum_{v=1}^{V}
    \mathcal{L}_{\mathrm{SIGReg}}
    \left(\{z_i^{(v)}\}_{i=1}^{B}\right).
\]
We use the LeJEPA SIGReg functional unchanged; the supplementary material
states the implementation and hyperparameters.
The intuition is simple: alignment preserves information shared across
different views of the same segment, while SIGReg keeps the representation
well spread for later linear probes. The longer geometry argument is placed in
the appendix because the paper's main question is scaling. We do not claim
that this objective preserves every downstream variable or is optimal for
every fMRI task.

With the pretraining method fixed, Section~\ref{sec:eval_protocol} defines the
pretraining data size, model size, training progress, and estimated compute
used below.

%% file: sections/04_protocol.tex
\section{Controlled Scaling Setup}
\label{sec:eval_protocol}

We hold the \neurojepa{} objective, ROI-time input, view construction,
architecture family, optimizer family, and downstream protocol fixed. We vary
pretraining data size, model size, and progress through the planned training
schedule.

\subsection{Data, Models, and Compute}

\begin{table*}[t]
\centering
\small
\setlength{\tabcolsep}{5pt}
\begin{tabularx}{\textwidth}{@{}X X r l@{}}
\toprule
\textbf{Dataset / source} & \textbf{Downstream task} &
\textbf{Subjects} & \textbf{Metrics} \\
\midrule
\multicolumn{4}{@{}l}{\textit{In-distribution (ID)}} \\
ABIDE~\citep{di2014autism} & Autism diagnosis (binary) & 871 &
Accuracy / macro-F1 \\
ABIDE & Age prediction & 871 & MSE / Pearson $r$ \\
PNC~\citep{satterthwaite2014neuroimaging} & Sex classification (binary) &
1,268 & Accuracy / macro-F1 \\
PPMI~\citep{marek2011parkinson} & Diagnosis classification (3-way) & 474 &
Accuracy / macro-F1 \\
HCP~\citep{van2013wu} & Sex classification (binary) & 1,010 &
Accuracy / macro-F1 \\
\midrule
\multicolumn{4}{@{}l}{\textit{Out-of-distribution (OOD)}} \\
ADNI~\citep{jack2008alzheimer} & Alzheimer's disease vs.\ control & 230 &
Accuracy / macro-F1 \\
ADNI\textsuperscript{$\dagger$} & Mild cognitive impairment vs.\ control & 292 &
Accuracy / macro-F1 \\
ADHD-200~\citep{adhd2012adhd} & ADHD diagnosis (binary) & 696 &
Accuracy / macro-F1 \\
BHRC~\citep{salum2025cohort} & Sex classification (binary) & 465 &
Accuracy / macro-F1 \\
NKI-RS~\citep{nooner2012nki} & Age prediction & 717 & MSE / Pearson $r$ \\
NKI-RS & Education classification (3-way) & 717 & Accuracy / macro-F1 \\
SALD~\citep{Wei2017StructuralAF} & Age prediction & 492 &
MSE / Pearson $r$ \\
\bottomrule
\end{tabularx}
\caption{\textbf{Downstream datasets and evaluation tasks.} Counts denote
unique participants after task-specific filtering and deduplication across the
training, validation, and test splits. Detailed task descriptions and metric
construction are provided in the supplementary material.}
\label{tab:scaling_tasks}
\end{table*}

The pretraining data contain fMRI signals from more than 200 source
datasets. To obtain the common ROI-time series, we spatially resample each fMRI
recording to 2\,mm isotropic resolution and temporally resample it onto a
common model-input grid with 0.72\,s spacing. We then use the 100-parcel
Schaefer cortical parcellation \citep{schaefer2018local} to extract a
100-channel cortical ROI-time series. All resulting ROI-time series undergo the
same temporal normalization.
The main crossed grid contains pretraining data sizes of 2,585, 3,810, 6,533,
11,673, 32,414, and 63,401 recordings. All levels use the same source datasets
and differ only in the participant fraction sampled within each source,
preserving source proportions up to rounding. They are subject-grouped and
strictly nested: participants enter with all recordings, and every smaller
level is contained in the next larger level.

Let $D$ denote \emph{pretraining data size}, measured by the number of
distinct recordings. Reusing a recording later in training does not increase
$D$. 

Let $N$ denote \emph{model size}, measured by the number of trainable encoder
parameters and excluding the projection head. The encoder is a Vision
Transformer over ROI-time patches, and the model family varies its width and
depth. The main scaling analyses cover model sizes from approximately 2
million to 92 million encoder parameters.
Each analysis reports its trained support rather than implying a complete
Cartesian grid. The supplementary material lists the architectures, and the
supplementary dataset table reports the recordings and participants from each
source.

Let $C$ denote cumulative training compute in EFLOP, where
$1\ \mathrm{EFLOP}=10^{18}$ floating-point operations. We estimate $C$ from
the dominant dense-matrix operations in the encoder,
the projection head, and the number of optimizer steps; the supplementary
material gives the accounting. For training run $i$, let $s$ be the
number of completed optimizer steps and $S_i$ its planned total. Training
progress is $p=s/S_i$, so $p=0.5$ is the midpoint of that run's schedule and
$p=1$ is its endpoint. Equal $p$ does not imply equal compute across runs.
For the allocation analysis in Section~\ref{sec:downstream_eval}, let $e$
denote \emph{epoch-equivalent exposure}: the cumulative number of recording
presentations divided by $D$. Thus, $e=1$ corresponds to one average pass over
the available pretraining recordings. Unlike $p$, which measures progress
relative to a run's planned schedule, $e$ measures how often the available
recordings have been presented.

\subsection{Evaluation Protocol}

The 12 fixed tasks span demographic, clinical, 
with both classification and regression. They cover sex, age,
education, autism, Parkinson's disease, Alzheimer's disease, mild cognitive
impairment, and ADHD across ABIDE, PNC, PPMI, SALD, ADNI, ADHD-200, BHRC, HCP,
and NKI-RS. Table~\ref{tab:scaling_tasks} lists the task groups, participant
counts, and metrics.
In Figure~\ref{fig:scaling}, classification and regression group tasks by
target type, while in-distribution (ID) and out-of-distribution (OOD) denote two fixed dataset groups listed in that
table.

Subjects follow 6:2:2 train, validation, and test partitions. All
recordings from one participant remain in one split. The train 
partitions remain fixed across five evaluation seeds.
The pretraining data exclude downstream validation and test participants. 
For each task, the encoder is frozen, and a linear probe is trained for each downstream task.
For cross-task plots, a classification score averages macro-F1. A regression score averages Pearson
correlation. 
Each score is oriented so that higher is better, averaged across the five
probe seeds. The plotted error is one minus score; the supplementary material
gives the exact construction.

%% file: sections/05_scaling.tex
\section{How Should fMRI Pretraining Scale?}
\label{sec:scaling}

\begin{figure*}[t]
    \centering
    \includegraphics[width=\textwidth]{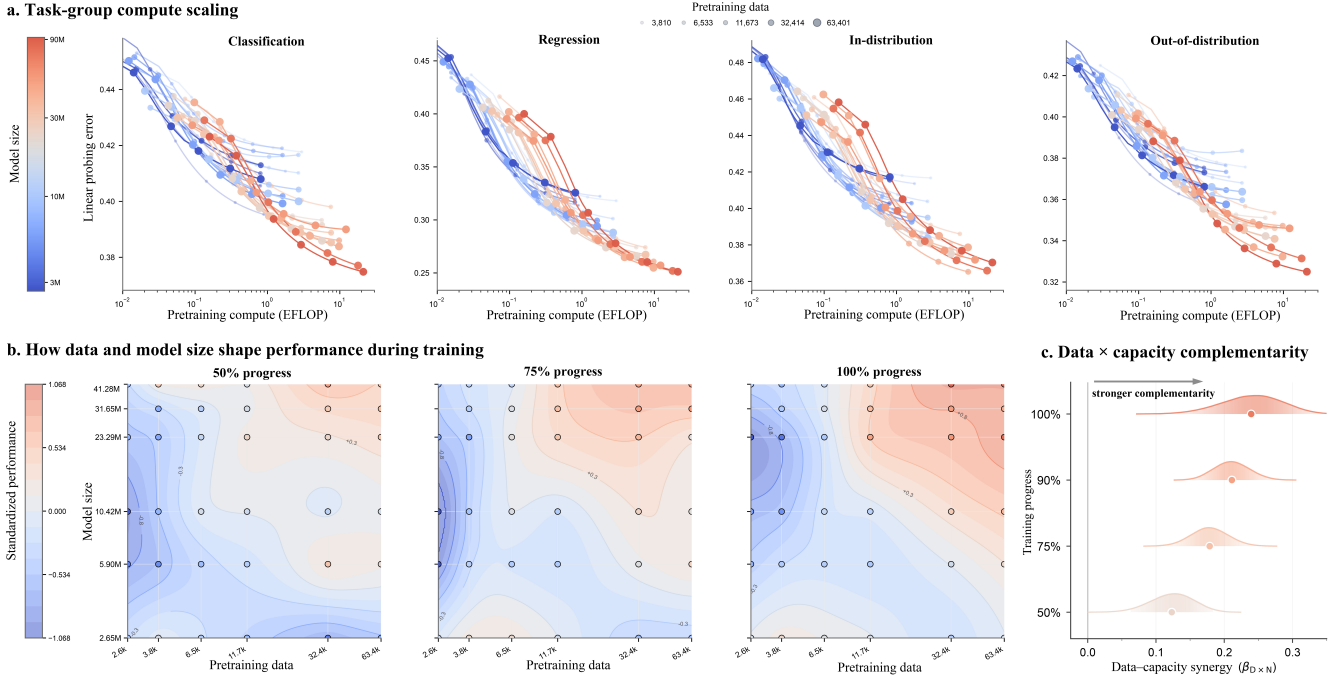}
    \caption{\textbf{Training trajectories and data--model-size
    relationships
    during training.}
    (a) Linear-probe error for fixed data--model configurations.
    (b) Standardized downstream scores at matched fractions of each training
    progress; circles are evaluated configurations and contours are
    descriptive interpolations. (c) Task-resampling uncertainty for
    $\beta_{D\times N}$; positive values mean that larger models show larger
    fitted gains from additional pretraining data.}
    \label{fig:scaling}
\end{figure*}

We organize the scaling analysis around training progress, the joint
data--model relationship, and comparisons at similar estimated compute. The
analyses use
higher-is-better, outer-validation scores. The data axis is
the nested pretraining data sequence defined in Section
\ref{sec:eval_protocol}. We describe this observed grid rather than
fit or extrapolate a universal scaling law.
Panel (a) shows the equivalent lower-is-better error; the remaining analyses
use the higher-is-better score.

\subsection{Compute Helps, but Scaling Choices Matter}

Figure~\ref{fig:scaling}(a) follows the task-group-averaged linear-probe error
of each fixed data--model configuration as training progresses. Across
classification, regression, ID, and OOD, error generally decreases as
estimated compute accumulates. The pattern is not strictly monotonic, but its
overall direction is consistent across all four task-group summaries.

At the upper end of the observed compute range, configurations combining larger
pretraining data sizes with larger model sizes increasingly occupy the
lower-error region. 
This is a descriptive pattern within the evaluated grid:
pretraining data size, model size, training progress, and
estimated compute change together, so Figure~\ref{fig:scaling}(a) does not
isolate the independent effect of any one factor.

Estimated compute nevertheless does not uniquely order the observed
configurations. Trajectories with overlapping or nearby compute ranges can
remain at different error levels and sometimes change their ordering during
training. Compute therefore tracks the amount of training but does not fully
describe how efficiently that compute is used. This motivates separating the
data-size and model-size axes and asking whether their relationships with
downstream performance depend on one another.

\subsection{Data Gains Depend on Model Size}

We next ask a concrete question: is additional pretraining data associated with
the same performance change for small and large models?
Figure~\ref{fig:scaling}(b) maps downstream performance over the observed
data--model grid at matched stages of training progress. If the two resources
were unrelated, moving toward more data would produce a similar fitted change
on every model-size row. Instead, the fitted relationship with data size becomes
more favorable in larger-model rows. Equivalently, the relationship with model
size becomes more favorable when more pretraining data are available.

To check whether this visual pattern is shared across tasks, we measure, for
each task, how the fitted gain from additional pretraining data changes as
model size increases. We average these changes over the 12 tasks and denote the
result by $\beta_{D\times N}$. A positive value has a direct interpretation:
larger models show larger fitted gains from additional pretraining data. A
value near zero would mean that the fitted data-size relationship is similar
across model sizes.

Figure~\ref{fig:scaling}(c) repeatedly resamples the task set to show the
uncertainty in $\beta_{D\times N}$. It is positive throughout the measured
training stages and larger at the end of training than at the midpoint. In
practical terms, increasing model size and increasing pretraining data size are
most effective when they are scaled together and trained sufficiently. This
result applies to the evaluated grid and does not establish a causal or
strictly monotonic relationship. The supplementary material provides the
regression definition, numerical
estimates, and sensitivity analyses.

\begin{figure}[t]
    \centering
    \includegraphics[width=\columnwidth]{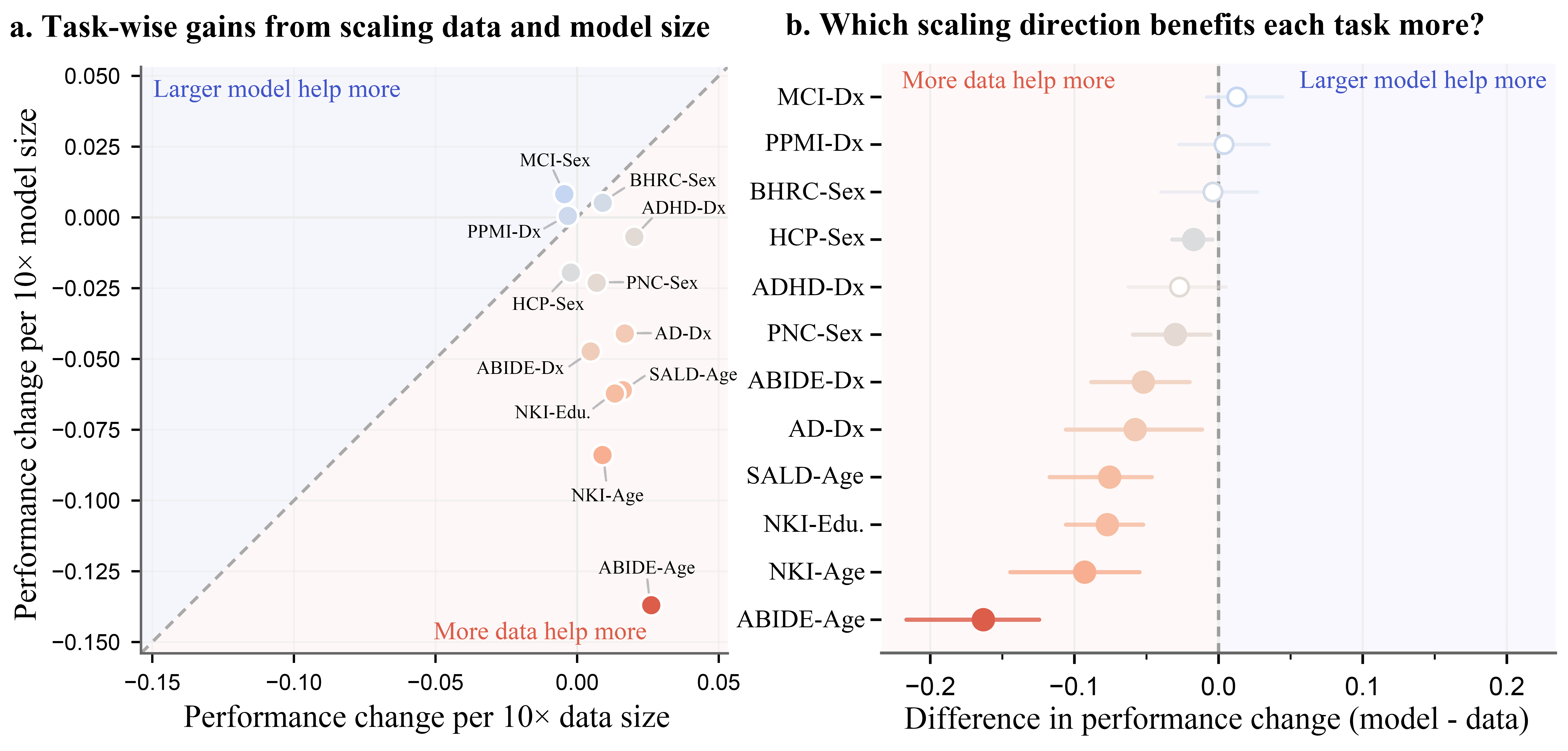}
    \caption{\textbf{Task-wise comparison of scaling pretraining data and
    model size at matched compute.}
    (a) Each task is positioned by the performance changes associated with
    increasing data size and model size; the diagonal denotes equal changes.
    (b) Their difference on one axis: values to the left favor more data,
    values to the right favor a larger model, and intervals crossing zero are
    unresolved.}
    \label{fig:matched_compute}
\end{figure}

\subsection{Most Tasks Favor Data at Matched Compute}

The previous analysis shows that pretraining data and model size work
together. In practice, however, the training budget is fixed. We therefore ask
which direction is associated with a larger downstream gain at the same
estimated compute: using more pretraining data or using a larger model.

Figure~\ref{fig:matched_compute}(a) compares these two directions for each
task. Moving to the right means that performance improves more as pretraining
data increase; moving upward means that performance improves more as model
size increases. Tasks below the diagonal show a larger gain from data scaling,
whereas tasks above it show a larger gain from model scaling.
Panel (b) places the same comparison on a single axis. Values to the left of
zero favor more data, values to the right favor a larger model, and an interval
crossing zero means that the current experiments cannot distinguish the two.

Eight of the twelve tasks favor more pretraining data, while the remaining
four are unresolved; none clearly favors a larger model. This does not mean
that model scaling never helps. Rather, within the evaluated compute range,
data scaling provides the more consistent improvement across downstream
tasks, whereas the benefit of a larger model is more task dependent. These
comparisons describe the observed grid rather than a universal allocation
rule. The supplementary material provides the
matched-compute construction and uncertainty analysis. Section
\ref{sec:downstream_eval} next asks whether these measurements can support a
budget-constrained allocation rule calibrated on ID tasks, and whether models
selected by that rule remain competitive on OOD tasks.

\FloatBarrier

%% file: sections/06_final_evaluation.tex
\section{OOD Evaluation of ID-Guided Models}
\label{sec:downstream_eval}

Section~\ref{sec:scaling} shows that additional compute generally helps, but
does not determine how that compute should be divided among pretraining data,
model size, and training duration. We now turn this observation into a
model-selection procedure. Given a compute budget, we use ID downstream
performance to choose a supported combination of pretraining data, model size,
and training duration. We then lock the selected model and evaluate it on OOD
tasks.

\paragraph{ID-guided model selection.}
Following \citet{muennighoff_data_constrained}, we fit an effective-data
response to the mean linear-probing performance on five ID tasks. The response
relates ID performance to pretraining data size, model size, and training
duration; together with the EFLOPs of each observed model, it ranks
configurations only within the measured range. Its form and parameters are
selected using complete-run ID validation and frozen before OOD evaluation.
The equations, formula comparisons, and diagnostic results are reported in
the supplementary material, together with an audit of all budget-eligible
\neurojepa{} models and data-heavy, model-heavy, longer-training, and
best-observed-ID rules using OOD rank and regret.

We refit this response using all eligible ID downstream results and
consider two prespecified compute budgets that cover two baseline-compute
regimes. At each budget, we select the observed combination of pretraining
data, model size, and training duration with the best predicted ID performance.
The two selected models and all evaluation settings are locked before OOD
evaluation.
The supplementary material reports validation that holds out complete
training runs or one data scale at a time.

The OOD datasets also appear in the descriptive scaling analyses in
Section~\ref{sec:scaling}. However, neither their validation nor test
performance enters the response fit, compute-budget definition, or model
selection in this section, and the OOD datasets are excluded from
\neurojepa{} pretraining. This is therefore a selection-held-out OOD
evaluation, not a claim that these datasets were absent from every descriptive
analysis. The exact fitting and compute-matching procedures are given in
the supplementary material.

\begin{figure}[t]
    \centering
    \includegraphics[width=\columnwidth]{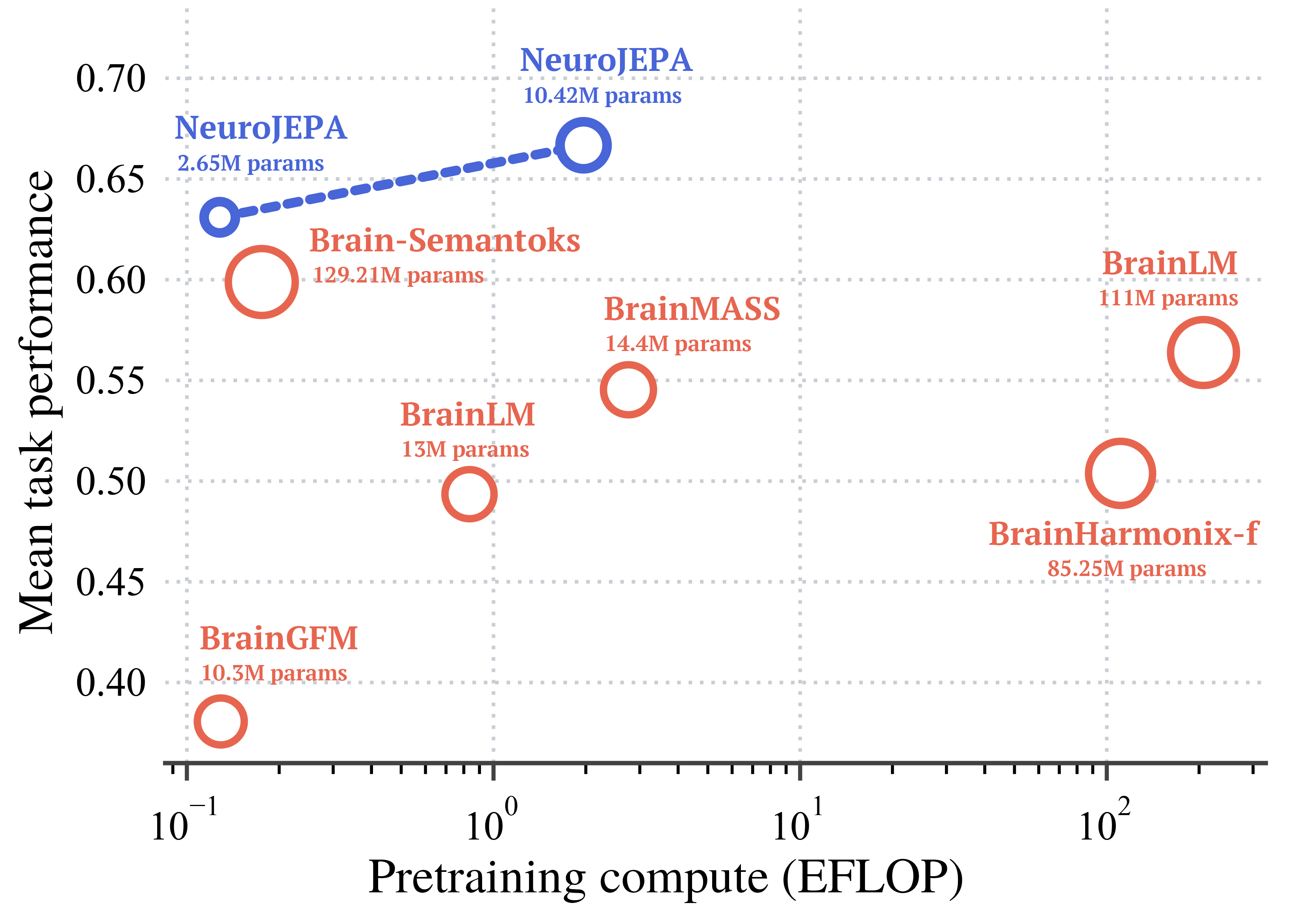}
    \caption{\textbf{OOD performance versus estimated pretraining compute.}
    Blue markers denote the two \neurojepa{} models selected using ID
    performance only, and red markers denote public fMRI
    foundation-model baselines; labels report model parameters. The vertical
    axis averages classification accuracy and Pearson correlation for
    regression across the OOD tasks in Table~\ref{tab:final_eval}.}
    \label{fig:ood_allocation}
\end{figure}

\input{sections/06_subject_tables}

\paragraph{Evaluation protocol.}
We evaluate the six OOD tasks reported in Table~\ref{tab:final_eval}: AD
classification, ADHD diagnosis,
BHRC sex classification, SALD age prediction, NKI age prediction, and NKI
education classification. Every encoder is frozen, and the same linear-probe
search is applied to all models. Classification tasks use
$L_2$-regularized logistic regression, whereas regression tasks use ridge
regression. Input features are standardized using statistics computed from
the training split. For regression, targets are also standardized using the
training-set mean and standard deviation. The regularization strength is
selected separately for every model, task, and run on the validation split,
using log loss for classification and mean-squared error for regression. The
selected probe, trained only on the training partition, is then evaluated once
on the corresponding test partition.

All splits are made at the subject level, with every recording from one
participant kept in the same split. The training partition contains 60\% of
the subjects and remains fixed. The original validation and test subjects are
combined into a 40\% holdout pool. Before each of five runs, this pool is
randomly divided in half using a different seed, producing a 60:20:20
train--validation--test split. The same seed-specific split is used for every
baseline and both \neurojepa{} configurations.

\paragraph{Baseline and compute accounting.}
Each fMRI foundation-model baseline is evaluated from its official released
weights using the accompanying code and required input preprocessing. For
every baseline and both selected \neurojepa{} models, we report the number of
trainable model parameters and estimated pretraining EFLOPs. Baseline EFLOPs
are reconstructed from the training details reported in the paper, released
code, configuration files, and checkpoint metadata. Because these values are
estimates, we report their assumptions and available ranges in
the supplementary material.

\paragraph{OOD results.}
Figure~\ref{fig:ood_allocation} shows that both ID-guided \neurojepa{}
models achieve a favorable OOD performance--compute trade-off, with the
10.42M model obtaining the highest average OOD performance among the compared
models. Table~\ref{tab:final_eval} shows that this result is not driven by a
single dataset: the 10.42M model achieves the best mean on at least one
reported metric in five of the six tasks, while the 2.65M model ranks second
on at least one metric in four tasks. Relative to the closest-compute baseline,
each selected model obtains a higher mean on five of the six task-wise metrics
summarized in Figure~\ref{fig:ood_allocation}. These results support ID-guided
allocation as a compute-efficient model-selection strategy, without implying
that one configuration is optimal for every OOD task.

\FloatBarrier

%% file: sections/06_subject_tables.tex
\begin{table*}[t]
\centering
\small
\setlength{\tabcolsep}{1mm}
\begin{tabular*}{\textwidth}{@{\extracolsep{\fill}} l r r *{6}{c}}
\toprule
\multirow{2}{*}{\textbf{Model}} & \multirow{2}{*}{\shortstack{\textbf{Model}\\\textbf{parameters}}} & \multirow{2}{*}{\shortstack{\textbf{Pretrain}\\\textbf{EFLOPs}}} & \multicolumn{2}{c}{\textbf{ADNI-AD}} & \multicolumn{2}{c}{\textbf{ADHD}} & \multicolumn{2}{c}{\textbf{NKI-Age}} \\
\cmidrule(lr){4-5}\cmidrule(lr){6-7}\cmidrule(lr){8-9}
 &  &  & Acc$\uparrow$ & F1$\uparrow$ & Acc$\uparrow$ & F1$\uparrow$ & MSE$\downarrow$ & $r\uparrow$ \\
\midrule
BrainLM-13M & 13M & 0.835 & 56.4 $\pm$ 3.6 & 56.4 $\pm$ 3.6 & \underline{58.9 $\pm$ 4.0} & 57.4 $\pm$ 4.7 & .937 $\pm$ .027 & .321 $\pm$ .015 \\
BrainLM-111M & 111M & 206.77 & \best{76.4 $\pm$ 3.0} & \best{76.4 $\pm$ 3.0} & 52.1 $\pm$ 2.8 & 50.4 $\pm$ 1.1 & .776 $\pm$ .056 & .503 $\pm$ .039 \\
BrainMASS & 14.4M & 2.755 & 60.4 $\pm$ 4.1 & 60.0 $\pm$ 4.0 & 55.9 $\pm$ 0.9 & 55.7 $\pm$ 1.1 & .816 $\pm$ .054 & .460 $\pm$ .047 \\
BrainGFM & 10.3M & 0.129 & 60.9 $\pm$ 3.0 & 60.7 $\pm$ 3.1 & 55.4 $\pm$ 1.6 & 52.7 $\pm$ 3.0 & .981 $\pm$ .026 & .248 $\pm$ .043 \\
BrainHarmonix-F & 85.25M & 110.90 & 69.8 $\pm$ 5.0 & 68.8 $\pm$ 5.5 & 53.4 $\pm$ 1.1 & 47.0 $\pm$ 8.3 & .878 $\pm$ .066 & .394 $\pm$ .048 \\
Brain-Semantoks & 129.21M & 0.1755 & 67.1 $\pm$ 6.5 & 67.1 $\pm$ 6.5 & 58.3 $\pm$ 1.94 & \underline{57.4 $\pm$ 1.94} & .741 $\pm$ .040 & .533 $\pm$ .031 \\
\midrule
\neurojepa{} & 2.65M & 0.128 & \underline{71.1 $\pm$ 4.2} & \underline{70.8 $\pm$ 4.5} & 54.52 $\pm$ 2.0 & 50.78 $\pm$ 7.4 & \underline{.573 $\pm$ .028} & \underline{.669 $\pm$ .034} \\
\neurojepa{} & 10.42M & 1.97 & 67.1 $\pm$ 3.8 & 66.7 $\pm$ 3.3 & \best{60.3 $\pm$ 1.0} & \best{60.1 $\pm$ 1.2} & \best{.438 $\pm$ .029} & \best{.761 $\pm$ .023} \\
\bottomrule
\end{tabular*}
\vspace{1.5mm}
\begin{tabular*}{\textwidth}{@{\extracolsep{\fill}} l r r *{6}{c}}
\toprule
\multirow{2}{*}{\textbf{Model}} & \multirow{2}{*}{\shortstack{\textbf{Model}\\\textbf{parameters}}} & \multirow{2}{*}{\shortstack{\textbf{Pretrain}\\\textbf{EFLOPs}}} & \multicolumn{2}{c}{\textbf{BHRC-Sex}} & \multicolumn{2}{c}{\textbf{NKI-Edu.}} & \multicolumn{2}{c}{\textbf{SALD-Age}} \\
\cmidrule(lr){4-5}\cmidrule(lr){6-7}\cmidrule(lr){8-9}
 &  &  & Acc$\uparrow$ & F1$\uparrow$ & Acc$\uparrow$ & F1$\uparrow$ & MSE$\downarrow$ & $r\uparrow$ \\
\midrule
BrainLM-13M & 13M & 0.835 & 58.0 $\pm$ 3.8 & 48.4 $\pm$ 8.9 & 52.5 $\pm$ 3.7 & 46.7 $\pm$ 4.3 & .915 $\pm$ .122 & .382 $\pm$ .071 \\
BrainLM-111M & 111M & 206.77 & 56.0 $\pm$ 5.9 & 44.2 $\pm$ 3.9 & 48.7 $\pm$ 4.0 & 46.2 $\pm$ 5.3 & .725 $\pm$ .044 & .548 $\pm$ .051 \\
BrainMASS & 14.4M & 2.755 & \best{61.1 $\pm$ 5.2} & 46.8 $\pm$ 6.4 & 56.2 $\pm$ 3.0 & 54.9 $\pm$ 4.6 & .781 $\pm$ .057 & .476 $\pm$ .036 \\
BrainGFM & 10.3M & 0.129 & 56.6 $\pm$ 5.5 & 48.9 $\pm$ 3.4 & 32.8 $\pm$ 4.7 & 26.2 $\pm$ 5.4 & 1.067 $\pm$ .067 & -.021 $\pm$ .054 \\
BrainHarmonix-F & 85.25M & 110.90 & 59.4 $\pm$ 4.3 & 49.2 $\pm$ 10.1 & 38.1 $\pm$ 4.5 & 36.9 $\pm$ 4.7 & .824 $\pm$ .100 & .422 $\pm$ .081 \\
Brain-Semantoks & 129.21M & 0.1755 & 58.9 $\pm$ 4.0 & \underline{53.2 $\pm$ 9.6} & 53.6 $\pm$ 8.0 & 52.0 $\pm$ 7.9 & .557 $\pm$ .037 & .681 $\pm$ .016 \\
\midrule
\neurojepa{} & 2.65M & 0.128 & 57.7 $\pm$ 3.3 & 45.9 $\pm$ 6.19 & \underline{57.7 $\pm$ 5.3} & \underline{55.8 $\pm$ 5.3} & \underline{.505 $\pm$ .035} & \underline{.706 $\pm$ .016} \\
\neurojepa{} & 10.42M & 1.97 & \underline{58.9 $\pm$ 3.1} & \best{56.2 $\pm$ 8.0} & \best{62.6 $\pm$ 4.83} & \best{62.2 $\pm$ 4.86} & \best{.428 $\pm$ .020} & \best{.762 $\pm$ .028} \\
\bottomrule
\end{tabular*}
\caption{Downstream evaluation with frozen linear probes. Classification tasks report accuracy and F1 in percent; Regression tasks report MSE and Pearson correlation. Each entry is the mean $\pm$ standard deviation across seeds. Best and second-best means are shown in bold and underlined, respectively.
ADNI-MCI results are reported in the supplementary material.}
\label{tab:final_eval}
\label{tab:external_eval}
\end{table*}

%% file: sections/07_discussion.tex
\section{Discussion and Limitations}

Estimated compute is a useful measure of training cost, but it does not
uniquely determine downstream performance. Performance generally improves as
training advances, yet configurations with similar EFLOPs can remain
separated. The fitted data-size association becomes more favorable as model
size increases and is stronger at the end of training. At matched compute, it
is larger than the model-size association for eight of twelve tasks; the other
four are unresolved. The configuration behind a budget therefore matters in
addition to its total compute.

The practical question is which combination of pretraining data size, model
size, and training progress makes the best use of fixed compute. Our results
motivate searching these choices jointly rather than treating a larger model
as a substitute for more data or training. Data-rich configurations are
important candidates within the evaluated range, but the results do not
prescribe a universal data--model ratio.

We make this decision concrete by fitting ID downstream performance over the
supported combinations. At two compute budgets, we select the combination
with the best predicted ID performance and lock it before OOD evaluation. The
selected models remain competitive across six OOD tasks, and the larger one
achieves the highest average OOD performance among the compared fMRI
foundation models. Thus, ID performance can guide a strong fixed-budget
choice, without showing that the selected combination is optimal for every
OOD task or beyond the evaluated candidates.

The main limitation concerns extrapolation of the fitted ID response, not the
relationships observed within the current grid. The six nested data sizes span
approximately 24.5-fold, with no substantially larger configuration reserved
for a prospective test. The response can compare configurations within or
close to this range but cannot predict far beyond it.
Holding source datasets and sampling proportions fixed strengthens the
comparison; a future corpus with new sites, scanners, or acquisition protocols
may require the numerical response to be fitted again.

Finally, we study one fixed method and ROI-time architecture family with frozen
linear probes. Task resampling omits independent pretraining-run variation, and
baseline EFLOPs are estimated. These limitations motivate larger held-out
scales and replicated key runs.

%% file: sections/08_appendix.tex
\twocolumn[
\begin{center}
  {\Large\bfseries Supplementary Material for\par}
  \vspace{2pt}
  {\LARGE\bfseries A Scaling Study for fMRI Foundation Models\par}
  \vspace{8pt}
\end{center}
\vspace{16pt}
]

\section{Supplementary Methods and Analyses}

\providecommand{\appfulltablesetup}{%
  \centering
  \scriptsize
  \setlength{\tabcolsep}{3.5pt}
  \renewcommand{\arraystretch}{1.02}}

\setlength{\textfloatsep}{8pt plus 2pt minus 2pt}
\setlength{\floatsep}{6pt plus 2pt minus 2pt}
\setlength{\dbltextfloatsep}{8pt plus 2pt minus 2pt}
\setlength{\dblfloatsep}{12pt plus 2pt minus 2pt}
\setcounter{dbltopnumber}{4}
\renewcommand{\dbltopfraction}{0.9}
\renewcommand{\textfraction}{0.05}
\captionsetup{skip=3pt}
\makeatletter
\setlength{\@dblfptop}{0pt}
\setlength{\@dblfpsep}{16pt}
\setlength{\@dblfpbot}{0pt plus 1fil}
\makeatother

This appendix follows the order of the main paper. It first records the fixed
pretraining and evaluation details needed for reproduction, then gives the
statistical definitions behind the scaling results, and finally documents the
ID-guided model-selection and OOD audits.

\subsection{Fixed Pretraining Method}
\label{app:eflops}

\subsubsection{Latent Geometry and SIGReg}
\label{app:objective_geometry}

The geometry argument provides intuition for the fixed objective; it is not a
new theorem and is not used to establish the scaling results. Let
$Z\in\mathbb{R}^{n\times d_{\mathrm{emb}}}$ contain $n$ centered frozen
embeddings and let $\mu_1,\ldots,\mu_r$ be the positive eigenvalues of
$\widehat{\Sigma}=Z^\top Z/n$. At fixed rank and total variance, distributing
variance evenly avoids directions with very little variation. In an idealized
linear model, this has two familiar consequences. For ridge regression with
penalty $\lambda_{\mathrm r}$ and a target vector of norm $\rho$, the largest
directional shrinkage is controlled by the smallest positive eigenvalue,
\[
  \sup_{\|\beta^\star\|_2=\rho}
  \left\|\mathbb{E}[\widehat{\beta}\mid Z]-\beta^\star\right\|_2^2
  =
  \frac{\lambda_{\mathrm r}^2\rho^2}
       {(\mu_{\min}+\lambda_{\mathrm r})^2}.
\]
Under homoscedastic noise with variance $\sigma^2$, the total ordinary
least-squares coefficient variance on the rank-$r$ representation subspace is
proportional to
\[
  \frac{\sigma^2}{n}\sum_{j=1}^{r}\frac{1}{\mu_j}.
\]
Equal positive eigenvalues minimize this quantity when their sum is fixed.
These calculations explain why a well-spread representation can be convenient
for unknown linear readouts and label-limited probes. They do not imply that
isotropy creates task information, improves every downstream task, or makes
the objective optimal for fMRI.

In \neurojepa{}, alignment encourages global and local views of the same
ROI-time segment to share a representation. SIGReg separately discourages
collapse and concentration into a few latent directions. We use the
Epps--Pulley SIGReg functional from LeJEPA~\citep{lejepa} unchanged. For a
batch of embeddings $\mathcal{Z}=\{z_i\}_{i=1}^{B}$ and $M$ random unit
directions $\mathcal{A}=\{a_m\}_{m=1}^{M}$ in
$\mathbb{R}^{d_z}$,
\[
  \mathcal{L}_{\mathrm{SIGReg}}(\mathcal{Z})
  =
  \frac{1}{M}\sum_{m=1}^{M}
  T_{\mathrm{EP}}\!\left(\{a_m^\top z_i\}_{i=1}^{B}\right),
\]
where $T_{\mathrm{EP}}$ compares the projected empirical distribution with a
standard Gaussian. We use $M=4096$ directions and 17 trapezoidal integration
points on $[-5,5]$. Directions are sampled from a standard Gaussian, normalized
to unit length, and synchronized across workers. Empirical characteristic
functions are averaged across workers before the statistic is evaluated.
SIGReg is computed separately for each of the eight view batches and then
averaged.

\subsubsection{View Construction}
\label{app:view_augmentation}

Let $\widetilde{x}_i\in\mathbb{R}^{R\times T_i^{\mathrm{raw}}}$ denote one recording. We first sample the segment
$x_i\in\mathbb{R}^{R\times T}$ used by the objective in
Section~\ref{sec:objective}. Two global views retain broad context and six
local views retain less context. Cropped views are resampled to fixed shapes,
perturbed, and independently z-scored over time within each ROI.
Table~\ref{tab:view_recipe} gives the complete recipe.

\begin{table*}[t]
\appfulltablesetup
\begin{tabular*}{\textwidth}{@{\extracolsep{\fill}}
    l c p{0.22\textwidth} p{0.23\textwidth}
    p{0.24\textwidth} c@{}}
\toprule
\textbf{View} & \textbf{Count} & \textbf{Temporal / ROI crop} &
\textbf{Noise and amplitude} & \textbf{Masking} & \textbf{Shift} \\
\midrule
Global 1 & 1 &
$70$--$100\%$ of each axis; resample to $100\times200$ &
Noise SD $0.6\%$ of each ROI's temporal SD; amplitude-scale SD $1.2\%$ &
$1.6\%$ ROI masking; with probability $0.04$, mask a contiguous
$5$--$10\%$ temporal span &
$\pm1$ \\
Global 2 & 1 &
$70$--$100\%$ of each axis; resample to $100\times200$ &
Noise SD $1.0\%$ of each ROI's temporal SD; amplitude-scale SD $2.0\%$ &
$2.0\%$ ROI masking; with probability $0.05$, mask a contiguous
$5$--$10\%$ temporal span &
$\pm2$ \\
Local & 6 &
$30$--$60\%$ of each axis; resample to $40\times80$ &
Noise SD $1.1\%$ of each ROI's temporal SD; amplitude-scale SD $2.2\%$ &
$2.4\%$ ROI masking; with probability $0.06$, mask a contiguous
$5$--$10\%$ temporal span &
$\pm2$ \\
\bottomrule
\end{tabular*}
\caption{\textbf{View construction and perturbation recipe.}
Shifts are circular and measured in time points. Before view construction,
temporal resampling is applied with probability $0.2$ using a scale sampled
from $[0.9,1.2]$ and linear interpolation.}
\label{tab:view_recipe}
\end{table*}

\subsubsection{Architecture and Optimization}
\label{app:implementation}

Table~\ref{tab:implementation_defaults} lists settings shared by all scaling
runs. View-specific settings are reported once in
Table~\ref{tab:view_recipe}.

\subsubsection{NeuroJEPA Compute Accounting}

We count the dominant dense matrix multiplications in the encoder and
projection head. One multiply--accumulate is two floating-point operations.
Normalization, activation functions, positional encoding, and objective
statistics are excluded. For view $v$ with spatial dimensions $(H_v,W_v)$,
patch size $(p_h,p_w)$, embedding dimension $d$, and $n_{\mathrm{reg}}$
register tokens, define
\[
  P_v=
  \left\lfloor\frac{H_v}{p_h}\right\rfloor
  \left\lfloor\frac{W_v}{p_w}\right\rfloor,
  \qquad
  T_v=P_v+1+n_{\mathrm{reg}}.
\]
Here $P_v$ is the number of patch tokens and $T_v$ also includes the class and
register tokens. With patch area $A=p_hp_w$, Transformer depth $L$, and MLP
ratio $r$, the forward cost of one view is
\[
  F_{\mathrm{view}}(v)
  =
  2P_vAd+
  L\!\left[(8+4r)T_vd^2+4T_v^2d\right].
\]
For a projection head
$d\rightarrow H_{\mathrm{proj}}\rightarrow H_{\mathrm{proj}}\rightarrow d_z$,
\[
  F_{\mathrm{proj}}
  =
  2\!\left(dH_{\mathrm{proj}}+H_{\mathrm{proj}}^2+
  H_{\mathrm{proj}}d_z\right).
\]
With $n_g=2$ global views and $n_l=6$ local views, we approximate backward
computation as twice the forward cost:
\[
  \begin{aligned}
  F_{\mathrm{train/sample}}
  =3\big[&
    n_gF_{\mathrm{view}}(g)+n_lF_{\mathrm{view}}(l)\\
    &+(n_g+n_l)F_{\mathrm{proj}}
  \big].
  \end{aligned}
\]
Let $K$ be the number of completed optimizer steps and
$B_{\mathrm{global}}$ the effective batch size. Cumulative compute is
\[
  \boxed{
  C(K)=
  \frac{K B_{\mathrm{global}}F_{\mathrm{train/sample}}}{10^{18}}
  \ \text{EFLOP}.
  }
\]
Thus, equal parameter counts need not imply equal compute: view shapes, batch
size, and the number of completed steps also enter the estimate.

\begin{table*}[t]
\appfulltablesetup
\begin{tabular*}{\textwidth}{@{\extracolsep{\fill}}
    p{0.23\textwidth}p{0.48\textwidth}p{0.20\textwidth}@{}}
\toprule
\textbf{Parameter} & \textbf{Meaning} & \textbf{Value} \\
\midrule
\multicolumn{3}{@{}l}{\textit{Encoder and projector}} \\
Input segment & Schaefer-100 cortical ROI-time matrix & $100\times200$ \\
Patch size & ROI-by-time patch size & $1\times40$ \\
Summary / register tokens & Learned non-patch tokens & $1 / 4$ \\
Encoder width and depth & Configuration-specific values &
Table~\ref{tab:model_configurations} \\
Transformer MLP ratio & Hidden-to-embedding width ratio & $4$ \\
Projection head & Shared MLP after the encoder &
$d_{\mathrm{emb}}\!\rightarrow\!2048\!\rightarrow\!2048\!\rightarrow\!512$ \\
Attention normalization & QKV bias and query/key normalization &
Bias enabled; LayerNorm \\
Residual regularization & Drop path and LayerScale initialization &
$0.1$; $10^{-4}$ \\
Position encoding & Learned positions plus RoPE &
rotary ratio $0.5$, base $300$ \\
\midrule
\multicolumn{3}{@{}l}{\textit{Objective}} \\
Loss weights & Alignment / SIGReg & $0.95 / 0.05$ \\
SIGReg settings & Hyperparameter, projections, integration points &
$\alpha_{\mathrm{SIG}}=0.5$; $4096$; $17$ \\
Regularized dimension & Projection dimension seen by SIGReg & $512$ \\
\midrule
\multicolumn{3}{@{}l}{\textit{Optimization}} \\
Optimizer & Decoupled weight decay & AdamW, $(0.9,0.999)$ \\
Learning rate & Reference rate and batch scaling &
$0.01\sqrt{B_{\mathrm{global}}/1024}$ \\
Schedule & Warm-up, cosine decay, terminal floor &
$10\%$; cosine; $10^{-6}$ \\
Weight decay & Initial value & $5\times10^{-4}$ \\
Layer-wise LR decay & Multiplicative decay across depth & $0.9$ \\
Gradient clipping / accumulation & Global norm; accumulation steps &
$3.0 / 1$ \\
\bottomrule
\end{tabular*}
\caption{\textbf{Shared implementation settings.}
The global batch size and number of optimizer steps are run specific.}
\label{tab:implementation_defaults}
\end{table*}

\begin{table*}[t]

\appfulltablesetup
\begin{tabular*}{\textwidth}{@{\extracolsep{\fill}}rrrrrr@{}}
\toprule
\textbf{Encoder parameters} &
\textbf{Depth} &
\textbf{Embedding dimension} &
\textbf{Heads} &
\textbf{Head dimension} &
\textbf{MLP hidden dimension} \\
\midrule
2.65M  & 12 & 128 & 2  & 64 & 512 \\
5.90M  & 12 & 192 & 3  & 64 & 768 \\
10.42M & 12 & 256 & 4  & 64 & 1024 \\
23.29M & 12 & 384 & 6  & 64 & 1536 \\
31.65M & 12 & 448 & 7  & 64 & 1792 \\
41.28M & 12 & 512 & 8  & 64 & 2048 \\
52.19M & 12 & 576 & 9  & 64 & 2304 \\
64.38M & 12 & 640 & 10 & 64 & 2560 \\
75.05M & 14 & 640 & 10 & 64 & 2560 \\
92.59M & 12 & 768 & 12 & 64 & 3072 \\
\bottomrule
\end{tabular*}
\caption{\textbf{Encoder configurations used in the scaling experiments.}
Parameter counts exclude the projection head.}
\label{tab:model_configurations}
\end{table*}

\subsubsection{Experimental Resource Use}
\label{app:resource_use}

The experiments used approximately 10,347 GPU-hours in total. This quantity
documents the computational resources consumed by the study; it is distinct
from the EFLOP estimates used for model comparison, which account for each
model's computation rather than hardware occupancy. Pretraining accounts for
most of the recorded GPU-hours, and most experiments were run on NVIDIA
A800-SXM4-80GB GPUs. All values below are rounded.

\begin{table*}[t]
\appfulltablesetup
\begin{minipage}[t]{0.48\textwidth}
\centering
\begin{tabular*}{\linewidth}{@{\extracolsep{\fill}}lrr@{}}
\toprule
\textbf{Experiment stage} & \textbf{GPU-hours} & \textbf{Share} \\
\midrule
Pretraining & $\sim$9,464 & 91.5\% \\
Downstream probing & $\sim$883 & 8.5\% \\
\midrule
Total & $\sim$10,347 & 100\% \\
\bottomrule
\end{tabular*}
\captionof{table}{\textbf{GPU-hours by experiment stage.}}
\label{tab:gpu_hours_stage}
\end{minipage}
\hfill
\begin{minipage}[t]{0.48\textwidth}
\centering
\begin{tabular*}{\linewidth}{@{\extracolsep{\fill}}lrr@{}}
\toprule
\textbf{GPU} & \textbf{GPU-hours} & \textbf{Share} \\
\midrule
NVIDIA A800-SXM4-80GB & $\sim$9,680 & 93.6\% \\
NVIDIA A100 & $\sim$667 & 6.4\% \\
\midrule
Total & $\sim$10,347 & 100\% \\
\bottomrule
\end{tabular*}
\captionof{table}{\textbf{GPU-hours by hardware type.}}
\label{tab:gpu_hours_hardware}
\end{minipage}

\vspace{8pt}

\begin{tabular*}{\textwidth}{@{\extracolsep{\fill}}rrrrr@{}}
\toprule
\textbf{Pretraining recordings} &
\textbf{Merged GPU-hours} &
\textbf{Downstream allocation} &
\textbf{Final GPU-hours} &
\textbf{Share} \\
\midrule
63,401 & 4,766.77 & 23.88 & 4,790.65 & 46.30\% \\
32,414 & 672.02 & 14.45 & 686.47 & 6.63\% \\
11,673 & 995.64 & 0.12 & 995.76 & 9.62\% \\
6,533 & 1,275.54 & 0.36 & 1,275.90 & 12.33\% \\
3,810 & 984.22 & 0.24 & 984.46 & 9.51\% \\
2,585 & 1,436.64 & 8.56 & 1,445.20 & 13.97\% \\
\midrule
Six-scale subtotal & 10,130.82 & 47.61 & 10,178.43 & 98.37\% \\
All-experiment total & 10,299.72 & 47.61 & 10,347.33 & 100\% \\
\bottomrule
\end{tabular*}
\captionof{table}{\textbf{GPU-hours attributed to each pretraining-data scale.}
The downstream-allocation column records the data-scale-specific downstream
cost assigned in the experiment ledger. Rows show the six nested data sizes
used in the main scaling analysis. The subtotal covers those six scales,
whereas the all-experiment total matches the stage and hardware totals above
and also includes experiments outside the main six-scale analysis. Small
discrepancies between displayed row sums and totals are due to rounding.}
\label{tab:gpu_hours_data_scale}
\end{table*}

\FloatBarrier

\subsection{Data and Downstream Evaluation}
\label{app:evaluation_details}

The full source catalogue, including participant and recording counts, appears
in Appendix~\ref{app:datasets}. The six pretraining data sizes in the main
analysis are subject-grouped, strictly nested subsets of that catalogue.
Downstream validation and test participants are excluded from pretraining.
For the final evaluation in Section~\ref{sec:downstream_eval}, the OOD datasets
are also excluded from the selected models' pretraining data. Their downstream
scores are used in the descriptive scaling analyses, but only the five ID tasks
are used to fit the response and select the two final models.

Classification summaries use macro-F1 and regression summaries use Pearson
correlation. Both are oriented so that higher is better. We first average each
task over five probe seeds and then give every task equal weight. Accuracy and
mean-squared error are reported as additional task-level metrics but do not
enter these cross-task summaries. In the interaction analyses below, task-wise
standardization is applied only to place regression coefficients from
different tasks on a common numerical scale.

\subsubsection{Frozen Linear Probes}
\label{app:probe_protocol}

For each frozen model, task, and evaluation seed, the feature standardizer is
fit only on the training partition. Classification uses $L_2$-regularized
logistic regression with inverse regularization strength
\[
  \kappa_{\mathrm{logit}}\in\{10^{-5},10^{-4},\ldots,10^{3}\},
\]
and regression uses ridge regression with
\[
  \lambda_{\mathrm{ridge}}\in\{10^{5},10^{4},\ldots,10^{-4}\}.
\]
Regression targets are standardized using the training-set mean and standard
deviation. Validation log loss selects $\kappa_{\mathrm{logit}}$, and validation
mean-squared error selects $\lambda_{\mathrm{ridge}}$. Logistic regression is optimized for at
most 5,000
iterations using the run seed. Exact ties follow the fixed grid order, which
favors stronger regularization. The training-fitted probe is evaluated
directly on the test partition and is not refit on the combined training and
validation data.

\begin{table*}[t]
\appfulltablesetup
\begin{tabular*}{\textwidth}{@{\extracolsep{\fill}}
    l p{0.29\textwidth} p{0.50\textwidth}@{}}
\toprule
\textbf{Dataset} & \textbf{Task} & \textbf{Label definition} \\
\midrule
ABIDE & Autism diagnosis & Autism spectrum disorder vs.\ typically developing control \\
ABIDE & Age prediction & Chronological age \\
PNC & Sex classification & Binary sex label \\
PPMI & Diagnosis classification & Prodromal Parkinson's disease, diagnosed Parkinson's disease, or control \\
HCP & Sex classification & Binary sex label \\
ADNI & AD diagnosis & Alzheimer's disease vs.\ cognitively normal control \\
ADNI & MCI diagnosis & Mild cognitive impairment vs.\ cognitively normal control \\
ADHD-200 & ADHD diagnosis & ADHD vs.\ typically developing control \\
BHRC & Sex classification & Binary sex label \\
NKI-RS & Age prediction & Chronological age \\
NKI-RS & Education classification & Primary (Grades 1--6), secondary (Grades 7--12), or higher education \\
SALD & Age prediction & Chronological age \\
\bottomrule
\end{tabular*}
\caption{\textbf{Downstream task definitions.}
Participant counts and reported metrics are given in
Table~\ref{tab:scaling_tasks}.}
\label{tab:task_definitions}
\end{table*}

\subsection{Evidence for the Scaling Findings}

\begin{figure*}[!t]
  \centering
  \includegraphics[width=\textwidth]{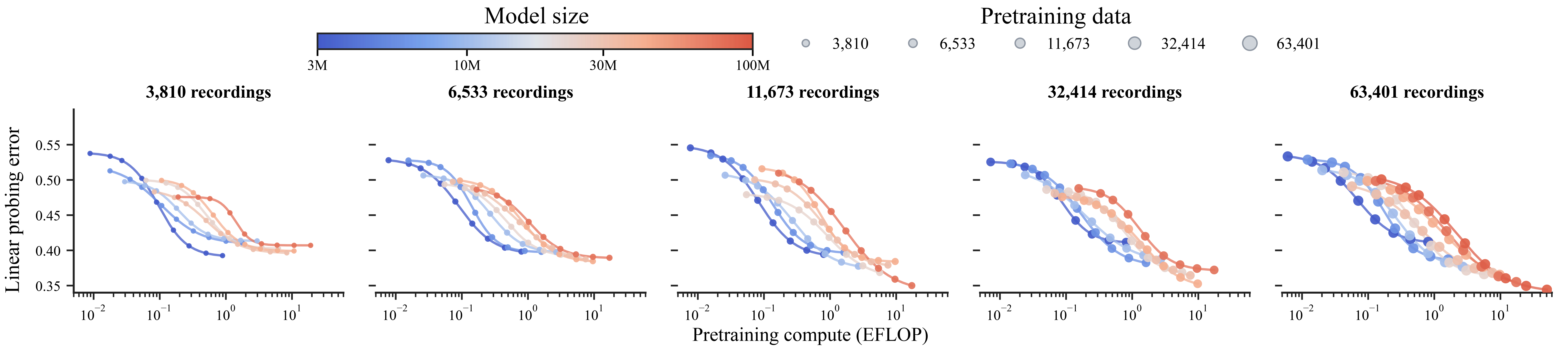}
  \caption{\textbf{Training trajectories grouped by pretraining data size.}
  Each panel fixes the number of pretraining recordings and compares model
  sizes as corrected pretraining compute increases. Lower linear-probing error
  is better.}
\label{fig:appendix_data_trajectories}
\end{figure*}

\begin{figure*}[!t]
  \centering
  \includegraphics[width=\textwidth]{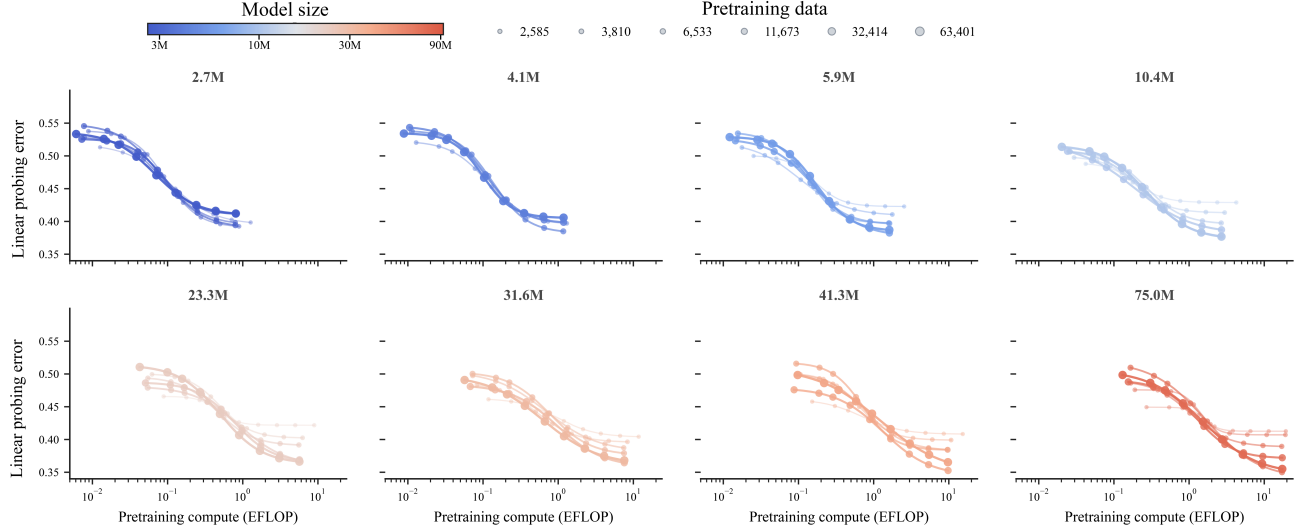}
  \caption{\textbf{Training trajectories grouped by model size.}
  Each panel fixes model size and compares pretraining data sizes as corrected
  pretraining compute increases. Lower linear-probing error is better.}
\label{fig:appendix_model_trajectories}
\end{figure*}

Figures~\ref{fig:appendix_data_trajectories} and
\ref{fig:appendix_model_trajectories} expand the compute trajectories in
Figure~\ref{fig:scaling}(a) along the two resource axes. The first groups
models by pretraining data size, while the second groups them by model size.

\subsubsection{How Data and Model Size Interact During Training}
\label{app:scheduled_interaction}

This analysis asks whether the fitted gain associated with additional
pretraining data changes with model size. Let $i$ index an observed
data--model configuration, $t$ a downstream task, and $p$ a measured fraction
of the configuration's planned training schedule. We use the validation score
for each downstream task; the same validation split is also used to choose the
probe regularization described above. The score is oriented so that higher is
better and standardized within task. Let $\widetilde d_i$ and
$\widetilde n_i$ denote standardized log data size and standardized log model
size. Their standardization is held fixed across $p$.

For every task and training fraction, we fit
\[
  \begin{aligned}
  s_{i,t}(p)
  ={}&a_t(p)+b_{D,t}(p)\widetilde d_i+b_{N,t}(p)\widetilde n_i\\
     &+\beta_{D\times N,t}(p)\widetilde d_i\widetilde n_i
     +\epsilon_{i,t}(p).
  \end{aligned}
\]
The coefficient $\beta_{D\times N,t}(p)$ answers the question directly: a
positive value means that the fitted data-size gain is larger for larger
models, or equivalently that the fitted model-size gain is larger at greater
data size. We give every task equal weight and define
\[
  \begin{aligned}
  \beta_{D\times N}(p)
  &=
  \frac{1}{12}\sum_{t=1}^{12}\beta_{D\times N,t}(p),\\
  \Delta\beta
  &=
  \beta_{D\times N}(1)-\beta_{D\times N}(0.5).
  \end{aligned}
\]

\begin{center}
\centering
\scriptsize
\setlength{\tabcolsep}{3pt}
\renewcommand{\arraystretch}{1.02}
\begin{tabular}{@{}ccp{0.43\columnwidth}@{}}
\toprule
\textbf{Training progress $p$} &
$\boldsymbol{\beta_{D\times N}(p)}$ &
\textbf{Interpretation} \\
\midrule
0.50 & 0.123 & Positive data--model-size relationship at schedule midpoint \\
0.75 & 0.179 & Positive relationship during later training \\
0.90 & 0.211 & Positive relationship near schedule completion \\
1.00 & 0.239 & Largest estimated relationship at completion \\
\bottomrule
\end{tabular}
\captionof{table}{\textbf{Data--model-size coefficient across training.}
The endpoint-minus-midpoint contrast is $\Delta\beta=0.116$, with paired
task-bootstrap 95\% interval $[0.006,0.216]$.}
\label{tab:scheduled_interaction}
\end{center}

The probability of a strictly increasing ordering across all four stages is
0.645. We therefore interpret the result as broad strengthening from mid to
late training, not as a strictly monotonic stage-by-stage increase. The paired
bootstrap resamples the 12 tasks as intact units and uses the same resampled
set at every $p$. Its interval measures sensitivity to the composition of the
fixed task panel; it does not represent variation across independent
pretraining runs or untrained configurations.

\subsubsection{Data versus Model Size at Matched Compute}
\label{app:matched_compute_responses}

This analysis asks which resource direction has the larger fitted relationship
with downstream performance when estimated compute is held fixed. We use 12
logarithmically spaced slices from 0.11 to 2.60 EFLOP. A trajectory contributes
to slice $c$ only when two saved models bracket that compute value in
$\log_{10}C$; values are linearly interpolated between those models and never
extrapolated. Each retained slice contains at least four model sizes, with at
least five data sizes represented for every included model size. The supported
cell count decreases from 34--35 at lower compute to 23 at the two highest
slices because fewer trajectories bracket those budgets.

At each task and compute slice, we fit
\[
  \begin{aligned}
  s_{i,t}(c)
  ={}&a_t(c)+\beta_{D,t}(c)\widetilde d_i(c)
      +\beta_{N,t}(c)\widetilde n_i(c)\\
     &+\eta_t(c)\widetilde d_i(c)\widetilde n_i(c)
      +\epsilon_{i,t}(c),
  \end{aligned}
\]
where $\widetilde d_i(c)$ and $\widetilde n_i(c)$ are the standardized log
data size and log model size among configurations available at slice $c$.
The main coefficients are averaged uniformly over the common compute grid:
\[
  \bar{\beta}_{D,t}
  =
  \frac{1}{|\mathcal{C}|}\sum_{c\in\mathcal{C}}\beta_{D,t}(c),
  \qquad
  \bar{\beta}_{N,t}
  =
  \frac{1}{|\mathcal{C}|}\sum_{c\in\mathcal{C}}\beta_{N,t}(c).
\]
Their contrast is
$\Delta_t=\bar{\beta}_{N,t}-\bar{\beta}_{D,t}$. A negative value means that
the fitted data-size relationship is larger; it does not by itself imply that
either relationship is positive.

We use 20,000 Bayesian-bootstrap replicates to test whether this contrast
depends on the represented data sizes, model sizes, and compute slices. Each
replicate reweights those three observed axes, refits the supported slices,
and recomputes $\Delta_t$. The resulting interval measures sensitivity to the
observed resource conditions. It does not include independent pretraining-run
or probe-seed uncertainty and is not adjusted for comparisons across tasks.

\begin{table*}[t]
\appfulltablesetup
\begin{tabular*}{\textwidth}{@{\extracolsep{\fill}}lrrrrl@{}}
\toprule
\textbf{Task} & $\boldsymbol{\bar{\beta}_{D,t}}$ &
$\boldsymbol{\bar{\beta}_{N,t}}$ &
$\boldsymbol{\Delta_t}$ & \textbf{95\% interval} & \textbf{Direction} \\
\midrule
ABIDE-Age & 0.026 & -0.137 & -0.163 & $[-0.217,-0.125]$ & Data size \\
NKI-Age   & 0.009 & -0.084 & -0.093 & $[-0.145,-0.055]$ & Data size \\
NKI-Edu.  & 0.016 & -0.061 & -0.077 & $[-0.106,-0.052]$ & Data size \\
SALD-Age  & 0.013 & -0.062 & -0.076 & $[-0.117,-0.046]$ & Data size \\
AD-Dx     & 0.017 & -0.041 & -0.058 & $[-0.106,-0.011]$ & Data size \\
ABIDE-Dx  & 0.005 & -0.047 & -0.052 & $[-0.088,-0.020]$ & Data size \\
PNC-Sex   & 0.007 & -0.023 & -0.030 & $[-0.059,-0.006]$ & Data size \\
HCP-Sex   & -0.002 & -0.020 & -0.017 & $[-0.032,-0.004]$ & Data size \\
ADHD-Dx   & 0.020 & -0.007 & -0.027 & $[-0.063,0.005]$ & Unresolved \\
BHRC-Sex  & 0.009 & 0.005  & -0.004 & $[-0.040,0.027]$ & Unresolved \\
PPMI-Dx   & -0.003 & 0.001 & 0.004  & $[-0.028,0.035]$ & Unresolved \\
MCI-Dx    & -0.005 & 0.008 & 0.013  & $[-0.008,0.044]$ & Unresolved \\
\bottomrule
\end{tabular*}
\caption{\textbf{Task-wise relationships at matched compute.}
The contrast is model size minus data size. Intervals below zero favor
pretraining data; intervals crossing zero are unresolved.}
\label{tab:matched_compute_tasks}
\end{table*}

As a sensitivity check, raising the lower end of the averaging range from
0.11 to 0.20 EFLOP leaves ABIDE-Age, ABIDE-Dx, AD-Dx, NKI-Age, NKI-Edu., and
SALD-Age data-size-favored. The other six intervals cross zero, and none
becomes model-size-favored.

\subsection{ID-Guided Selection under Fixed Compute}
\label{app:ood_selection}

The response is fitted to the equal-weight mean validation performance across
five ID tasks: ABIDE age, ABIDE diagnosis, HCP sex, PNC sex, and PPMI
diagnosis. The OOD tasks are included in the descriptive scaling analyses in
Section~\ref{sec:scaling}, but none of their scores enters the response fit,
compute budgets, or model selection. The response follows the effective-data family of
\citet{muennighoff_data_constrained}, with coefficients re-estimated for fMRI
downstream error and compute taken from Appendix~\ref{app:eflops}.

\subsubsection{Response, Fitting, and Formula Comparison}

For each saved model, we average the five ID scores and use one minus this
average as ID error. A trailing five-model median is applied within each
training run. Because it uses only the current and preceding saved models, it
does not leak later performance into an earlier compute budget. Runs receive
equal total weight so that densely saved trajectories do not dominate.

Let $D$ be pretraining data size, $N$ model size, and $e$ epoch-equivalent
exposure. The number of repeated passes beyond the first is
$R_D=e-1$. Following Eq.~14 of
\citet{muennighoff_data_constrained}, repeated exposure is represented by
\[
  D'
  =
  D\left[
  1+R_D^\star\left(1-\exp(-R_D/R_D^\star)\right)
  \right],
\]
where the positive parameter $R_D^\star$ controls how quickly the added value
of repeated passes decreases. The model size supported by the fitted
single-pass relation is defined using positive scale coefficients $A_N$ and
$B_D$ and positive exponents $\alpha_N$ and $\beta_D$:
\[
  N_{\mathrm{opt}}(D)
  =
  \left[
  \frac{\alpha_N A_N}{\beta_D B_D}D^{\beta_D}
  \right]^{1/\alpha_N}.
\]
Defining
\[
  U_N=\min\{N_{\mathrm{opt}}(D),N\},
  \quad
  R_N=\max\{N/U_N-1,0\},
\]
the effective model size is
\[
  N'
  =
  U_N\left[
  1+R_N^\star\left(1-\exp(-R_N/R_N^\star)\right)
  \right].
\]
The fitted ID response is
\[
  \widehat E_{\mathrm{ID}}(D,N,e)
  =
  E_\infty+\frac{A_N}{(N')^{\alpha_N}}
  +\frac{B_D}{(D')^{\beta_D}},
\]
where $E_\infty$ is the empirical error floor and $R_N^\star$ is the positive
capacity-decay parameter in the definition of $N'$. This response separates
pretraining data, model size, and repeated exposure, but is used only to
compare models inside the observed range. Let $C_i$ be the estimated
pretraining compute of model $i$. At budget $C_0$, the
selection rule is
\[
  i^\star(C_0)
  =
  \underset{i:\,C_i\le C_0}{\arg\min}\;
  \widehat E_{\mathrm{ID}}(D_i,N_i,e_i).
\]

The fit contains 42 complete training runs, 4,125 saved models, six data
sizes, and seven model sizes. We divide $D$ and $N$ by their geometric means
(10,750.24 recordings and 16.561 million parameters) for numerical
conditioning. All seven positive parameters are optimized in log space with
a run-balanced Huber objective and multistart L-BFGS-B. The fMRI grid does not
contain a complete single-pass experiment: the earliest evaluated state of a
run occurs after 3--154 epoch-equivalent exposures. The result is therefore a
joint within-grid fit of the response family, not a replication of the
original single-pass fitting protocol.

\begin{center}
\scriptsize
\setlength{\tabcolsep}{1.8pt}
\renewcommand{\arraystretch}{1.0}
\resizebox{\columnwidth}{!}{%
\begin{tabular}{@{}lrrrrr@{}}
\toprule
\textbf{Response} & \textbf{Run RMSE} & \textbf{Run regret} &
\textbf{Hold-$D$ RMSE} & \textbf{Hold-$N$ RMSE} & \textbf{Tail RMSE} \\
\midrule
Eq.~14: data and capacity decay & 0.0130 & 0.0080 & 0.0131 & 0.0135 & 0.0127 \\
Data decay only & 0.0134 & 0.0080 & -- & -- & -- \\
Capacity decay only & 0.0140 & 0.0068 & -- & -- & -- \\
No repeat decay & 0.0195 & 0.0202 & 0.0217 & 0.0190 & 0.0219 \\
\bottomrule
\end{tabular}}
\captionof{table}{\textbf{Response comparison and grouped holdouts.}
Complete training runs, data sizes, or model sizes are removed as units.
Tail prediction fits the early portion of each run and evaluates its later
models. Regret is the ID-score gap from the best observed model at matched
compute.}
\label{tab:response_validation}
\end{center}

On complete-run holdout, Eq.~14 reduces RMSE by 33.4\% relative to the model
that treats every repeated pass as new data and improves all six grouped folds
($p=0.0156$, one-sided paired Wilcoxon test). Most of the improvement comes
from modeling the diminishing value of repeated data passes. Adding the
capacity-decay term to the data-only response changes RMSE by only 2.9\%.
Run-cluster bootstrap estimates place $R_N^\star$ at its upper search bound in
73\% of samples, so the present model range does not identify a finite
capacity-decay scale. Raw versus smoothed trajectories and run-balanced versus
unweighted objectives give similar RMSE and data-decay estimates. These
diagnostics support the repeated-data correction but do not justify
interpreting either decay scale as a universal number of useful passes or
capacity multiples.

\subsubsection{Selected Models}

After grouped validation, the response is refit on all eligible ID results.
Under the strict constraints $C\le0.128$ and $C\le1.97$ EFLOP, it selects the
models in Table~\ref{tab:selected_budget_allocations}. Both budgets use the
same selection rule; ties are resolved in favor of lower compute. The fitted
response and observed smoothed ID performance identify the same model at each
budget.

\begin{strip}
\appfulltablesetup
\begin{tabular*}{\textwidth}{@{\extracolsep{\fill}}
    l r r r r r r r r@{}}
\toprule
\textbf{Budget} & \textbf{$D$} & \textbf{$N$} &
\textbf{Epoch} & \textbf{Progress} & \textbf{EFLOP} &
\textbf{Budget used} & \textbf{Pred.\ ID} & \textbf{Obs.\ ID} \\
\midrule
0.128 & 63,401 & 2.654M & 63 & 0.158 &
0.1276 & 99.7\% & 0.5449 & 0.5420 \\
1.970 & 63,401 & 10.417M & 291 & 0.729 &
1.9658 & 99.8\% & 0.6091 & 0.6108 \\
\bottomrule
\end{tabular*}
\captionof{table}{\textbf{Models selected at the two compute budgets.}
$D$ is pretraining data size and $N$ is model size. Progress is the number of
completed optimizer steps divided by the planned steps; because each listed
run uses a fixed number of steps per epoch, this is also the selected epoch
divided by the planned endpoint. Predicted and observed ID columns are
higher-is-better scores.}
\label{tab:selected_budget_allocations}
\end{strip}

\subsubsection{Two-Budget OOD Audit}

A training run is eligible at a budget if it contains at least one saved model
whose estimated compute does not exceed that budget. Within each eligible run,
we retain the saved model with the highest predicted ID score; exact ties favor
lower compute. The audit score is the mean OOD test score across five
evaluation runs, with equal weight given to AD diagnosis, ADHD diagnosis, BHRC
sex, ADNI-MCI diagnosis, NKI age, and NKI education. Classification uses
macro-F1 and regression uses Pearson correlation. Regret is the difference
from the best eligible audit score at the same budget.

This archived six-task audit is a candidate-level diagnostic and differs from
the external benchmark aggregate in Figure~\ref{fig:ood_allocation}, which includes
SALD age instead of ADNI-MCI and uses classification accuracy. Neither audit
enters the response fit or the selection of the two final models.

\begin{center}
\scriptsize
\setlength{\tabcolsep}{2.4pt}
\renewcommand{\arraystretch}{1.02}
\begin{tabular}{@{}lrrrr@{}}
\toprule
\textbf{Budget (EFLOP)} & \textbf{Eligible models} &
\textbf{Audit score} & \textbf{Audit rank} & \textbf{Audit regret} \\
\midrule
0.128 & 35 & 0.5791 & 8  & 0.0211 \\
1.970 & 42 & 0.6240 & 24 & 0.0359 \\
\bottomrule
\end{tabular}
\captionof{table}{\textbf{OOD audit of the two ID-selected models.}
Rank and regret are computed among the eligible run-level candidates at the
same compute budget. Higher score, lower rank, and lower regret are better.}
\label{tab:two_budget_ood_audit}
\end{center}

The compact audit above summarizes the available eligible-candidate
comparison. Separate numerical outputs for the
data-heavy, model-heavy, longer-training, and best-observed-ID heuristic rules
were not retained, so we do not report unverified rule-specific values here.

\subsubsection{Additional ADNI-MCI Results}

ADNI-MCI uses the same frozen-probe protocol as the other OOD classification
tasks. Its full baseline comparison is reported here, rather than in the main
evaluation table, solely because of the main-paper page limit.

\begin{center}
\scriptsize
\setlength{\tabcolsep}{2.4pt}
\renewcommand{\arraystretch}{1.02}
\begin{tabular}{@{}lrrrr@{}}
\toprule
\textbf{Model} & \textbf{Parameters} & \textbf{Pretrain EFLOPs} &
\textbf{Accuracy} & \textbf{Macro-F1} \\
\midrule
BrainLM-13M       & 13M     & 0.835  & $50.7\pm6.0$ & $50.7\pm6.0$ \\
BrainLM-111M      & 111M    & 206.77 & $57.6\pm5.2$ & $57.5\pm5.1$ \\
BrainMASS         & 14.4M   & 2.755  & $53.8\pm3.0$ & $53.4\pm2.9$ \\
BrainGFM          & 10.3M   & 0.129  & $51.7\pm7.7$ & $49.3\pm6.4$ \\
BrainHarmonix-F   & 85.25M  & 110.90 & $57.2\pm4.3$ & $57.2\pm4.2$ \\
Brain-Semantoks   & 129.21M & 0.1755 & $59.7\pm2.1$ & $59.4\pm2.2$ \\
\midrule
\neurojepa{}      & 2.65M   & 0.128  & $57.2\pm3.0$ & $57.2\pm2.9$ \\
\neurojepa{}      & 10.42M  & 1.97   & $54.8\pm6.7$ & $54.7\pm7.0$ \\
\bottomrule
\end{tabular}
\captionof{table}{\textbf{ADNI-MCI frozen-probe results.}
Classification metrics are percentages. Entries are mean $\pm$ standard
deviation across five evaluation seeds; baseline results use official released
weights.}
\label{tab:selected_mci_results}
\end{center}

\subsubsection{Baseline Compute Reconstruction}

For each released baseline, we instantiate the official architecture with its
pretraining input shape and measure the training FLOPs of one example. Let
$F_{\mathrm{ex}}$ be this cost, including forward and backward computation,
$B$ the global batch size, and $K$ the number of optimizer steps. When the
release reports $K$, compute is
\[
  C_{\mathrm{base}}=F_{\mathrm{ex}}BK/10^{18}.
\]
When it instead reports $n$ examples and $E$ epochs, we reconstruct
$K=E\lfloor n/B\rfloor$. An official cumulative FLOP counter takes precedence
over either reconstruction. Checkpoint metadata take precedence over
configuration files, released code, and paper descriptions; parameter count
alone is not used as a compute estimate.

\begin{center}
\scriptsize
\setlength{\tabcolsep}{2.4pt}
\renewcommand{\arraystretch}{1.03}
\begin{tabularx}{\columnwidth}{@{}l >{\raggedright\arraybackslash}X@{}}
\toprule
\textbf{Model} & \textbf{Compute reconstruction} \\
\midrule
BrainLM-13M & $1.0037$ TFLOPs/example $\times 832{,}000$ examples
$=0.835$ EFLOP. \\
BrainLM-111M & Official cumulative counter: $206.77$ EFLOP. \\
BrainMASS & $21.349$ GFLOPs/example $\times 129{,}024{,}000$ examples
$=2.755$ EFLOP. \\
BrainGFM & $32.3295$ GFLOPs/example $\times 4{,}000{,}000$ graph samples
$=0.129$ EFLOP. \\
BrainHarmonix-F & $4.3848$ TFLOPs/example $\times 25{,}292{,}800$ examples
$=110.90$ EFLOP. \\
Brain-Semantoks & $46.953$ GFLOPs/example $\times 3{,}737{,}600$ examples
$=0.1755$ EFLOP. \\
\bottomrule
\end{tabularx}
\captionof{table}{\textbf{Reconstructed baseline pretraining compute.}
BrainLM-111M uses its official cumulative counter.}
\label{tab:baseline_compute_reconstruction}
\end{center}

Schedules and metadata follow the official releases. These reconstructions do
not imply identical hardware utilization or implementation efficiency. The
strict model-selection budgets permit no compute overrun. Available training
descriptions determine the precision of each baseline estimate, and unresolved
pretraining-cohort overlap is treated as a limitation rather than evidence of
compute equivalence.

\FloatBarrier

%% file: sections/09_baselines.tex
\section{Baseline Models}

The public baselines span ROI sequences, functional connectomes,
atlas-derived graphs, and multimodal inputs. We use each model's official
weights and required preprocessing.

\paragraph{BrainLM.}
BrainLM~\citep{brainlm} learns from ROI-level time series by predicting masked
regional activity from visible context.

\paragraph{BrainMASS.}
BrainMASS~\citep{brainmass} represents BOLD recordings as functional
connectomes and aligns augmented brain networks during masked-ROI pretraining.

\paragraph{BrainGFM.}
BrainGFM~\citep{braingfm} learns from atlas-derived brain graphs using
contrastive learning, masked graph autoencoding, and learned prompts.

\paragraph{BrainHarmonix.}
BrainHarmonix~\citep{brainharmonix} separately pretrains structural and
functional MRI before fusing them through shared brain-hub tokens.

\paragraph{Brain-Semantoks.}
Brain-Semantoks~\citep{brain_semantoks} combines semantic tokenization and
self-distillation to model regional fMRI dynamics.

%% file: sections/09_datasets.tex
\section{Dataset Catalogue}
\label{app:datasets}
Table~\ref{tab:datasets} lists the 201 curated sources. We retain collections
with resting-state fMRI and remove task-only datasets. Counts follow
dataset-level curation, and one participant may contribute multiple
recordings. Downstream validation and test participants, together with the OOD
datasets in Section~\ref{sec:downstream_eval}, are excluded from the two final
models' pretraining data. The rows remain here for documentation and link to
the primary publication or repository record.

\begingroup
\fontsize{8.5}{9.1}\selectfont
\setlength{\tabcolsep}{4pt}
\renewcommand{\arraystretch}{1.16}
\setlength{\LTpre}{3pt}
\setlength{\LTpost}{0pt}
\begin{longtable}{@{}r >{\raggedright\arraybackslash}p{0.70\textwidth} r r@{}}
\caption{\textbf{Curated fMRI sources.} Counts precede the nested pretraining
subsets and downstream exclusions.}\label{tab:datasets}\\
\toprule
\textbf{ID} & \textbf{Dataset and reference} & \textbf{Subjects} & \textbf{Recordings} \\
\midrule
\endfirsthead
\caption[]{\textbf{Curated fMRI sources} (continued).}\\
\toprule
\textbf{ID} & \textbf{Dataset and reference} & \textbf{Subjects} & \textbf{Recordings} \\
\midrule
\endhead
\midrule
\multicolumn{4}{r}{\footnotesize Continued on the next page}\\
\endfoot
\bottomrule
\endlastfoot
1 & ABCD~\citep{casey2018adolescent} & 8101 & 8101 \\
2 & ABIDE~\citep{di2014autism} & 871 & 871 \\
3 & ADHD-200~\citep{adhd2012adhd} & 696 & 1061 \\
4 & ADNI~\citep{jack2008alzheimer} & 497 & 497 \\
5 & AOMIC (PIOP1)~\citep{snoek2021amsterdam} & 206 & 206 \\
6 & AOMIC (PIOP2)~\citep{snoek2021amsterdam} & 224 & 224 \\
7 & BHRC~\citep{salum2025cohort} & 465 & 465 \\
8 & Caltech Conte Center~\citep{kliemann2022caltech} & 102 & 356 \\
9 & CCNP~\citep{poldrack2016phenome} & 193 & 381 \\
10 & CHCP~\citep{ge2023increasing} & 304 & 304 \\
11 & CoRR~\citep{zuo2014open} & 434 & 1016 \\
12 & Emo-FilM~\citep{morgenroth2025emofilm} & 9 & 65 \\
13 & FCON~\citep{biswal2010toward} & 48 & 48 \\
14 & HBN~\citep{alexander2017open} & 1342 & 1342 \\
15 & HCP~\citep{van2013wu} & 1011 & 4044 \\
16 & ISYB~\citep{Gao2022ACM} & 187 & 187 \\
17 & MDD~\citep{yan2019reduced} & 3525 & 3525 \\
18 & NKI~\citep{nooner2012nki} & 717 & 717 \\
19 & PNC~\citep{satterthwaite2014neuroimaging} & 1268 & 1268 \\
20 & PPMI~\citep{marek2011parkinson} & 474 & 474 \\
21 & SALD~\citep{Wei2017StructuralAF} & 493 & 493 \\
22 & SLIM~\citep{liu2017longitudinal} & 133 & 256 \\
23 & 7T Resting-state~\citep{zhang2025rest7t} & 89 & 267 \\
24 & ABIDE II~\citep{dimartino2017abide2} & 1050 & 1412 \\
25 & AHDC~\citep{openneuro_ds005896} & 262 & 2451 \\
26 & Aging MultiEcho~\citep{spreng2022aging} & 298 & 1788 \\
27 & ALS-FTD UHF~\citep{openneuro_ds007036} & 33 & 33 \\
28 & AOMIC (ID1000)~\citep{snoek2021amsterdam} & 881 & 881 \\
29 & Anxiety CBT Rest~\citep{openneuro_ds006045} & 83 & 423 \\
30 & Aphasia Recovery Cohort~\citep{openneuro_ds004884} & 192 & 508 \\
31 & Bilingual Rest~\citep{openneuro_ds001747} & 92 & 92 \\
32 & Bilingualism Brain~\citep{openneuro_ds001796} & 64 & 64 \\
33 & BOLD Variability During Cognitive Control~\citep{openneuro_ds005270} & 158 & 158 \\
34 & BTC Preop~\citep{openneuro_ds001226} & 36 & 36 \\
35 & C-PRO~\citep{ito2022cpro} & 96 & 864 \\
36 & CO2 Resting~\citep{openneuro_ds005418} & 35 & 35 \\
37 & Cognitive Control Theoretic Mechanisms~\citep{openneuro_ds003831} & 73 & 73 \\
38 & Cognitive Training~\citep{kable2017cogtrain} & 42 & 69 \\
39 & Complex Multi-Echo~\citep{ingrova2026complexmultiecho} & 83 & 2988 \\
40 & Dystonia Rest~\citep{openneuro_ds006395} & 90 & 90 \\
41 & Emotional Learning Rest~\citep{openneuro_ds004109} & 30 & 87 \\
42 & Experience Sampling~\citep{openneuro_ds004134} & 36 & 36 \\
43 & Food Brain Study~\citep{openneuro_ds004697} & 82 & 82 \\
44 & GRACE~\citep{openneuro_ds007694} & 137 & 137 \\
45 & Gut-Brain Axis~\citep{openneuro_ds004648} & 88 & 88 \\
46 & Head Motion Intervention~\citep{openneuro_ds000256} & 24 & 168 \\
47 & Hearing Loss Connectome~\citep{openneuro_ds005026} & 82 & 82 \\
48 & Hemodynamic Timing~\citep{openneuro_ds004645} & 15 & 30 \\
49 & Hippocampal Memory Rest~\citep{openneuro_ds007063} & 52 & 52 \\
50 & HRV Biofeedback~\citep{openneuro_ds003357} & 32 & 64 \\
51 & HRV Biofeedback (ds003823)~\citep{openneuro_ds003823} & 176 & 978 \\
52 & Human Es-fMRI~\citep{openneuro_ds002799} & 20 & 80 \\
53 & Inhibitory Control Youth~\citep{openneuro_ds005754} & 124 & 372 \\
54 & Insulin Menstrual Rest~\citep{openneuro_ds006893} & 15 & 60 \\
55 & IQSEC2~\citep{openneuro_ds003468} & 26 & 634 \\
56 & LEMON~\citep{babayan2018lemon} & 226 & 226 \\
57 & Large-scale Pre/Post-surgical Patients~\citep{openneuro_ds005713} & 234 & 234 \\
58 & MBSR~\citep{openneuro_ds005016} & 147 & 347 \\
59 & MND~\citep{openneuro_ds005874} & 59 & 236 \\
60 & Monash RsPETMR~\citep{openneuro_ds002898} & 27 & 162 \\
61 & MSC~\citep{gordon2017msc} & 10 & 100 \\
62 & Multi-echo Cambridge~\citep{openneuro_ds000258} & 89 & 355 \\
63 & Multiband Acceleration~\citep{openneuro_ds003540} & 32 & 316 \\
64 & MyConnectome~\citep{poldrack2015myconnectome} & 1 & 132 \\
65 & Neurocon~\citep{conp_neurocon} & 43 & 85 \\
66 & NeuroEmo~\citep{abgeena2025neuroemo} & 40 & 80 \\
67 & NIMH Healthy Research Volunteer Dataset~\citep{openneuro_ds005752} & 219 & 984 \\
68 & Olfactory Meningioma~\citep{openneuro_ds007345} & 56 & 56 \\
69 & Oxytocin~\citep{openneuro_ds004725} & 87 & 87 \\
70 & Pediatric Anxiety Rest~\citep{openneuro_ds006303} & 149 & 546 \\
71 & Penn LEAD~\citep{openneuro_ds007116} & 127 & 414 \\
72 & Precision Aging Network~\citep{openneuro_ds007522} & 697 & 697 \\
73 & Prototype/Exemplar Aging~\citep{openneuro_ds006686} & 64 & 64 \\
74 & QTAB~\citep{strike2023qtab} & 413 & 2014 \\
75 & QTIM~\citep{openneuro_ds004169} & 1195 & 2453 \\
76 & RT Disengagement~\citep{openneuro_ds006683} & 59 & 59 \\
77 & Sleep EEG-fMRI~\citep{gu2023sleepeegfmri} & 33 & 255 \\
78 & Sleepy Brain~\citep{openneuro_ds000201} & 85 & 165 \\
79 & SONG~\citep{openneuro_ds004592} & 27 & 54 \\
80 & SRPBS Multi-disorder MRI~\citep{tanaka2021srpbs} & 1020 & 1020 \\
81 & SUDMEX-CONN~\citep{angelesvaldez2022sudmex} & 141 & 141 \\
82 & Synaesthesia~\citep{racey2023synaesthesia} & 127 & 508 \\
83 & T1 Diabetes EF Rest~\citep{openneuro_ds006156} & 64 & 64 \\
84 & Tao Wu Parkinson's Dataset~\citep{wu2017taowu} & 40 & 40 \\
85 & Temporal Lobe Epilepsy - UNAM~\citep{openneuro_ds004469} & 65 & 65 \\
86 & THINGS-fMRI~\citep{hebart2023things} & 3 & 354 \\
87 & Tinnitus~\citep{openneuro_ds002896} & 38 & 38 \\
88 & TMS-fMRI~\citep{openneuro_ds005498} & 148 & 151 \\
89 & Transdiagnostic Connectome Project~\citep{chopra2025tcp} & 241 & 1613 \\
90 & Tumor Patients Task/Rest~\citep{openneuro_ds005003} & 42 & 156 \\
91 & VASO Pulsatility Rest~\citep{openneuro_ds006212} & 23 & 142 \\
92 & Wakayama PsyRS~\citep{nitrc_wakayama_psyrs} & 200 & 583 \\
93 & WashU120~\citep{openneuro_ds000243} & 120 & 203 \\
94 & Yale NeuroConnect~\citep{openneuro_ds007286} & 410 & 815 \\
95 & Yale Pupillometry~\citep{openneuro_ds003673} & 27 & 54 \\
96 & Yale Reading~\citep{openneuro_ds005339} & 89 & 143 \\
97 & DMT-HAR-MED~\citep{openneuro_ds006644} & 40 & 80 \\
98 & 5.0T Visual Scene~\citep{openneuro_ds007354} & 20 & 680 \\
99 & Frontoparietal Plasticity~\citep{openneuro_ds003849} & 92 & 184 \\
100 & Habit Learning~\citep{openneuro_ds004299} & 123 & 1356 \\
101 & NIMH CAT-D~\citep{openneuro_ds004627} & 130 & 2369 \\
102 & NIMH Ketamine Mechanism of Action Study~\citep{openneuro_ds005917} & 58 & 1471 \\
103 & NIMH METeR~\citep{openneuro_ds004787} & 5 & 148 \\
104 & NODEAP~\citep{openneuro_ds006693} & 48 & 996 \\
105 & Infra-Low Frequency Neurofeedback~\citep{openneuro_ds004101} & 9 & 18 \\
106 & ON-Harmony~\citep{openneuro_ds004712} & 20 & 184 \\
107 & PAFIN~\citep{openneuro_ds006131} & 51 & 1020 \\
108 & PLP NF1~\citep{openneuro_ds003027} & 18 & 715 \\
109 & PSYCH-REST~\citep{openneuro_ds006952} & 34 & 271 \\
110 & Pragmatic Language~\citep{openneuro_ds003481} & 145 & 572 \\
111 & PsiConnect~\citep{openneuro_ds006110} & 65 & 2020 \\
112 & RPN Signature Study 1~\citep{openneuro_ds002608} & 41 & 41 \\
113 & RPN Signature Study 2~\citep{openneuro_ds002609} & 49 & 49 \\
114 & Reading Brain Project L1 Adults~\citep{openneuro_ds003974} & 52 & 312 \\
115 & Reading Brain Project L2 Adults~\citep{openneuro_ds003988} & 56 & 336 \\
116 & Reinforcement-Learning Generalization~\citep{openneuro_ds005230} & 44 & 130 \\
117 & SUDMEX-TMS~\citep{openneuro_ds003037} & 53 & 154 \\
118 & Speech Disfluencies~\citep{openneuro_ds003469} & 81 & 81 \\
119 & SpiDa-MRI~\citep{openneuro_ds004630} & 49 & 343 \\
120 & Temporal Dynamics of Emotional Music~\citep{openneuro_ds003085} & 39 & 156 \\
121 & Truecrime~\citep{openneuro_ds004965} & 133 & 133 \\
122 & Udall Pilot ANT~\citep{openneuro_ds001907} & 9 & 126 \\
123 & Valenced Tactile Information~\citep{openneuro_ds005449} & 115 & 1830 \\
124 & Visuomotor Rotation Adaptation Experiment~\citep{openneuro_ds004021} & 32 & 320 \\
125 & Visuomotor Rotation Learning and Reward-based Motor Learning~\citep{openneuro_ds005598} & 45 & 308 \\
126 & Resting State and Arithmetic Task~\citep{openneuro_ds002422} & 46 & 138 \\
127 & Fibromyalgia Emotion Regulation Dataset~\citep{openneuro_ds004144} & 66 & 131 \\
128 & Closed-eyes Depression and Healthy Controls~\citep{openneuro_ds002748} & 72 & 72 \\
129 & HC / PD-NC / PD-MCI Resting-State MRI~\citep{openneuro_ds005892} & 55 & 55 \\
130 & rewardBeast~\citep{openneuro_ds002738} & 35 & 121 \\
131 & MULTI-CLARID~\citep{openneuro_ds005795} & 34 & 39 \\
132 & Modafinil alters intrinsic functional connectivity of the right posterior insula: a pharmacological resting state fMRI study~\citep{openneuro_ds000133} & 26 & 156 \\
133 & Brain connectivity predicts placebo response across chronic pain clinical trials~\citep{openneuro_ds000208} & 76 & 76 \\
134 & Multi-echo fMRI replication sample of autobiographical memory, prospection and theory of mind reasoning tasks~\citep{openneuro_ds000210} & 31 & 93 \\
135 & Multiband Multi-Echo Imaging of Simultaneous Oxygenation and Flow Timeseries for Resting State Connectivity~\citep{openneuro_ds000216} & 7 & 28 \\
136 & Cost Analysis TBI~\citep{openneuro_ds000220} & 26 & 66 \\
137 & MPI-Leipzig\_Mind-Brain-Body~\citep{openneuro_ds000221} & 316 & 994 \\
138 & ds000245\_R1.0.0~\citep{openneuro_ds000245} & 45 & 45 \\
139 & rsfMRI\_single\_session\_EEG\_NF~\citep{openneuro_ds001408} & 52 & 104 \\
140 & Human hippocampal replay during rest prioritizes weakly-learned information and predicts memory performance~\citep{openneuro_ds001454} & 24 & 48 \\
141 & Layer VASO in visual system~\citep{openneuro_ds001547} & 4 & 8 \\
142 & 100 runs at 3T~\citep{openneuro_ds001553} & 3 & 19 \\
143 & InterTVA. A multimodal MRI dataset for the study of inter-individual differences in voice perception and identification.~\citep{openneuro_ds001771} & 40 & 40 \\
144 & The physiological effects of non-invasive brain stimulation fundamentally differ across the human cortex~\citep{openneuro_ds001927} & 23 & 133 \\
145 & Functional Connectivity of Music-Induced Analgesia in Fibromyalgia~\citep{openneuro_ds001928} & 40 & 160 \\
146 & Auditory localization with 7T fMRI~\citep{openneuro_ds001942} & 9 & 9 \\
147 & Multi-domain task battery (MDTB)~\citep{openneuro_ds002105} & 18 & 37 \\
148 & Simultaneous eeg-fmri for a speeded discrimination task with confidence~\citep{openneuro_ds002158} & 20 & 20 \\
149 & Caltech rsfMRI Dataset~\citep{openneuro_ds002232} & 6 & 12 \\
150 & Dense Investigation of Variability of Affect (DIVA)~\citep{openneuro_ds002278} & 3 & 152 \\
151 & Neuroimaging predictors of creativity in healthy adults~\citep{openneuro_ds002330} & 66 & 132 \\
152 & Yale\_Single\_Subject\_Task\_Rest30x~\citep{openneuro_ds002372} & 1 & 53 \\
153 & Meditacion Interocepcion~\citep{openneuro_ds002614} & 1 & 1 \\
154 & 28andMe~\citep{openneuro_ds002674} & 1 & 59 \\
155 & Headmold~\citep{openneuro_ds002735} & 11 & 44 \\
156 & YanDataBIDS~\citep{openneuro_ds002750} & 3 & 3 \\
157 & Cast-induced plasticity~\citep{openneuro_ds002766} & 3 & 197 \\
158 & Aging~\citep{openneuro_ds002872} & 39 & 39 \\
159 & SoccerCAN~\citep{openneuro_ds002940} & 24 & 46 \\
160 & Two sessions of resting state with closed eyes for patients with depression in treatment course (NFB, CBT or No treatment groups)~\citep{openneuro_ds003007} & 29 & 58 \\
161 & Social Processes Initiative in Neurobiology of the Schizophrenia(s) Traveling Human Phantoms~\citep{openneuro_ds003011} & 4 & 30 \\
162 & How ovarian hormones influence the behaviroal activation and inhibition system through the dopamine pathway~\citep{openneuro_ds003114} & 49 & 49 \\
163 & Using anesthesia-induced loss of consciousness to identify biomarkers of conscious awareness in the healthy human brain~\citep{openneuro_ds003171} & 17 & 68 \\
164 & Monash vis-fPET-fMRI~\citep{openneuro_ds003382} & 10 & 30 \\
165 & Brain Network Mechanisms of Visual Shape Completion~\citep{openneuro_ds003404} & 19 & 19 \\
166 & Identification of an Amygdala-Thalamic Circuit That Acts as a Central Gain Mechanism in Taste Perception~\citep{openneuro_ds003424} & 28 & 28 \\
167 & Protecting the Aging Brain - Diet-Study~\citep{openneuro_ds003437} & 12 & 36 \\
168 & Protecting the Aging Brain, Case-Study~\citep{openneuro_ds003455} & 1 & 2 \\
169 & Emotion Category and Face Perception Task Optimized for Multivariate Pattern Analysis~\citep{openneuro_ds003548} & 16 & 16 \\
170 & Reward biases spontaneous neural reactivation during sleep~\citep{openneuro_ds003574} & 18 & 18 \\
171 & ScanTrain~\citep{openneuro_ds003659} & 3 & 130 \\
172 & Parallel systems for social and spatial reasoning~\citep{openneuro_ds003814} & 10 & 300 \\
173 & PE-Update~\citep{openneuro_ds003835} & 24 & 46 \\
174 & Resting-state for 34 younger and 28 older adults~\citep{openneuro_ds003871} & 62 & 62 \\
175 & Pre-Post rehabilitation fMRI data of post-stroke patients.~\citep{openneuro_ds003999} & 29 & 58 \\
176 & rest\_eye~\citep{openneuro_ds004158} & 20 & 40 \\
177 & Perinatal Stroke~\citep{openneuro_ds004498} & 1 & 64 \\
178 & Ironia VEV~\citep{openneuro_ds004533} & 41 & 41 \\
179 & CS-DSI~\citep{openneuro_ds004737} & 20 & 20 \\
180 & language fMRI~\citep{openneuro_ds004765} & 71 & 71 \\
181 & POP~\citep{openneuro_ds005017} & 39 & 40 \\
182 & The Neural Basis of Visual Shape Completion in Schizophrenia and Bipolar Disorder~\citep{openneuro_ds005073} & 30 & 30 \\
183 & Neural Correlates of Lidocaine Analgesic (NLA) Study~\citep{openneuro_ds005088} & 27 & 54 \\
184 & 28andHe~\citep{openneuro_ds005115} & 1 & 40 \\
185 & Weill Cornell Medicine Multi-echo (WCM-ME) Dataset~\citep{openneuro_ds005118} & 1 & 220 \\
186 & AMRI 16-N-0031 sleep1~\citep{openneuro_ds005127} & 13 & 21 \\
187 & SoCal Kinesia and Incentivization for Parkinson's Disease (SKIP): Ultra-High Field Functional Connectivity~\citep{openneuro_ds005264} & 28 & 84 \\
188 & BABA: Naturalistic fMRI and MEG Dataset~\citep{openneuro_ds005346} & 29 & 29 \\
189 & Protecting the Aging Brain - fMRI study of the brain in ketosis~\citep{openneuro_ds005405} & 101 & 404 \\
190 & Priority~\citep{openneuro_ds005464} & 29 & 87 \\
191 & Circadian misalignment and energy balance~\citep{openneuro_ds005525} & 11 & 176 \\
192 & A comparison of resting state functional magnetic resonance imaging to invasive electrocortical stimulation for sensorimotor mapping in pediatric patients~\citep{openneuro_ds005573} & 16 & 32 \\
193 & Mitchell\_Hacker\_2013~\citep{openneuro_ds005603} & 8 & 53 \\
194 & The DBS-fMRI dataset~\citep{openneuro_ds005849} & 14 & 880 \\
195 & China's Social Fake News database release with brain structural, functional, and behavioural measures~\citep{openneuro_ds005875} & 43 & 43 \\
196 & QNL NegativeBOLD Database~\citep{openneuro_ds006148} & 255 & 290 \\
197 & Chicago Attention and Thoughts~\citep{openneuro_ds006515} & 60 & 231 \\
198 & Night Owls Scan Club~\citep{openneuro_ds006707} & 4 & 184 \\
199 & Linking Subjective Experience of Anxiety to Brain Function using Natural Language Processing.~\citep{openneuro_ds006948} & 75 & 447 \\
200 & Dense longitudinal single-subject multimodal MRI dataset acquired via self-administered scanning~\citep{openneuro_ds007328} & 1 & 458 \\
201 & Multi-scale, multi-modal imaging assessment of trajectories of cognitive impairment in Multiple Sclerosis~\citep{openneuro_ds007908} & 28 & 270 \\
\end{longtable}
\endgroup